\pdfoutput=1

\documentclass[11pt]{article}

\usepackage{acl}

\usepackage{times}
\usepackage{latexsym}

\usepackage[T1]{fontenc}

\usepackage[utf8]{inputenc}

\usepackage{microtype}

\usepackage{inconsolata}

\usepackage{graphicx}
\usepackage{colortbl}
\usepackage{enumitem}
\usepackage{array}
\usepackage{mathrsfs}
\usepackage{amsthm}
\usepackage{amsmath}
\usepackage{amsfonts}
\usepackage[noend]{algpseudocode}
\usepackage{multirow}
\usepackage{booktabs}
\usepackage{color}
\usepackage{arydshln}
\usepackage{makecell}
\usepackage{soul}
\usepackage{cleveref}
\usepackage{bbding}
\usepackage{subfigure} 
\usepackage{colortbl}
\usepackage{subcaption}

\renewcommand{\thefootnote}{\fnsymbol{footnote}}

\title{Benchmarking and Improving Large Vision-Language Models for Fundamental Visual Graph Understanding and Reasoning}

\author{Yingjie Zhu\textsuperscript{\rm 1,2}, Xuefeng Bai\textsuperscript{\rm 1}\footnotemark[1], Kehai Chen\textsuperscript{\rm 1,2}, Yang Xiang\textsuperscript{\rm 2}\footnotemark[1], Jun Yu\textsuperscript{\rm 3}, Min Zhang\textsuperscript{\rm 1,2} \\
     \textsuperscript{\rm 1}Institute of Computing and Intelligence, Harbin Institute of Technology, Shenzhen, China\\
     \textsuperscript{\rm 2}Peng Cheng Laboratory, Shenzhen, China\\
     \textsuperscript{\rm 3}School of Intelligence Science and
Engineering, Harbin Institute of Technology, Shenzhen, China\\
    \texttt{\{baixuefeng,chenkehai,yujun,zhangmin2021\}@hit.edu.cn} \\
    \texttt{zhuyj@stu.hit.edu.cn},  \texttt{xiangy@pcl.ac.cn}\\
    }

\begin{document}
\maketitle
\footnotetext[1]{Corresponding Author}
\renewcommand{\thefootnote}{\arabic{footnote}}
\begin{abstract}
Large Vision-Language Models (LVLMs) have demonstrated remarkable performance across diverse tasks. 
Despite great success, recent studies show that LVLMs encounter substantial limitations when engaging with visual graphs. 
To study the reason behind these limitations,
we propose \textsc{VGCure}, a comprehensive benchmark covering 22 tasks for examining the fundamental graph understanding and reasoning capacities of LVLMs. 
Extensive evaluations conducted on 14 LVLMs reveal that LVLMs are weak in basic graph understanding and reasoning tasks, particularly those concerning relational or structurally complex information.
Based on this observation, we propose a structure-aware fine-tuning framework to enhance LVLMs with structure learning abilities through three self-supervised learning tasks. 
Experiments validate the effectiveness of our method in improving LVLMs' performance on fundamental and downstream graph learning tasks, as well as enhancing their robustness against complex visual graphs.\footnote{Our dataset and code are available at \url{https://github.com/AAAndy-Zhu/VGCure}.}
\end{abstract}

\section{Introduction}


Graphs serve as a fundamental data structure across a wide range of domains, including social network analysis \cite{10.1145/3485447.3512194}, recommendation systems \cite{zhang-etal-2023-sentiment-analysis}, knowledge graphs \cite{zhang-etal-2024-question}, chemistry~\cite{cao-etal-2024-presto}, biomedical molecules~\cite{liu-etal-2023-molca} and
semantic reasoning \citep{bai-etal-2022-graph}.
Existing methods have achieved great success in enhancing understanding and reasoning abilities in graph-based tasks~\cite{kim-etal-2023-kg,chen2024llaga}.
However, these approaches typically focus on specific graph types or tasks, posing challenges in designing versatile systems that are suitable for various tasks and graphs across diverse domains.

\begin{figure}[t]
    \centering
    \includegraphics[width=0.9\linewidth]{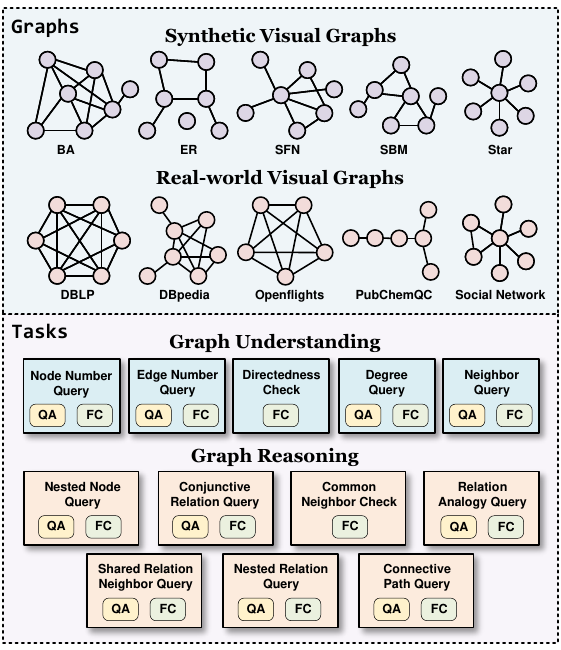}
    \vspace{-0.2cm}
    \caption{Overall of our \textsc{VGCure} benchmark.}
    \label{fig:introduction}
\end{figure}


\begin{table*}[t]
   \centering
    \resizebox{0.92\linewidth}{!}{
\begin{tabular}{lccccc}
\toprule
\textbf{Benchmark}   & \textbf{Evaluation Topic}  & \textbf{\# Tasks} & \textbf{Graph Type}     & \textbf{Anonymity} & \textbf{\# Scale} \\
\midrule
\textbf{VisionGraph} & Graph Theory Problems  & 10      & Synthetic                                              & \textcolor{green}{\Checkmark}               & 3,000           \\
\textbf{\citet{ai-etal-2024-advancement}}  & Multi-hop Reasoning   & N/A    & Real-world                                              & \textcolor{red}{\XSolidBrush}                & 1,355             \\
\textbf{GVLQA}  & Graph Theory Problems                & 7       & Synthetic                                & \textcolor{green}{\Checkmark}                & 157,896           \\
\hdashline
\textit{our} \textbf{\textsc{VGCure}}  & Fundamental Graph Understanding and Reasoning     & 22               & Synthetic \& Real-world              & \textcolor{green}{\Checkmark}               & 223,646\\
\bottomrule
\end{tabular}}
 \vspace{-0.2cm}
\caption{Comparisons among different visual graph analysis benchmarks for LVLMs, where \textit{\# Tasks} and \textit{\# Scale} represent the number of task types and test samples, respectively.}
\label{tab:introduction}
\end{table*}


Recently, Large Vision-Language Models (LVLMs) have exhibited outstanding performance across a wide range of downstream tasks by unifying various inputs in the form of images and processing them with human-like understanding and reasoning abilities~\cite{9961954, zheng-etal-2024-multimodal, zhang2025evaluating}. 
This triggers a growing interest in employing LVLMs for graph learning problems, as the vision modality offers a natural and intuitive way for comprehending structural information and facilitating general graph reasoning~\cite{pmlr-v162-poklukar22a}. 
However, recent studies~\cite{wei2024gita,li2024visiongraph,ai-etal-2024-advancement} reveal significant challenges for LVLMs in graph-based learning tasks, 
where LVLMs achieve less than 15\% accuracy on mathematical graph reasoning tasks, markedly below their performance in image and text reasoning~\cite{li2024visiongraph}. 
Therefore, a natural challenge arises: 
\textit{why do LVLMs fail in graph learning, and how to enhance LVLMs to process graphs like professionals?}


To address this challenge, we begin by identifying a research gap in existing evaluations that limit our understanding of LVLMs. 
Current work primarily focuses on graph theory problems or multi-hop reasoning tasks~\cite{wei2024gita,ai-etal-2024-advancement}, which are  complex and require a combination of diverse cognitive abilities. 
Thus it remains uncertain whether LVLMs have strong \textbf{fundamental understanding and reasoning capabilities} in processing visual graphs---such as recognizing the basic components of a graph and making basic logical inferences based on graph data---which is crucial for determining whether the limitations stem from a lack of fundamental abilities or other higher-order capabilities.


To this end, we present the \textbf{V}ision \textbf{G}raph \textbf{C}omprehensive \textbf{U}nderstanding and \textbf{RE}asoning benchmark, \textbf{\textsc{VGCure}}, designed to thoroughly evaluate the fundamental understanding and reasoning capabilities of LVLMs on visual graphs.
As shown in Fig.\ref{fig:introduction}, \textsc{VGCure} evaluates the fundamental capabilities of LVLMs diverse challenges, including 9 graph understanding tasks and 13 graph reasoning tasks.
Moreover, \textsc{VGCure} features 10 anonymized graph structures from both synthetic and real-world sources, offering a robust testbed for assessing LVLMs' proficiency in handling diverse graphs. 
Experiments on 14 representative LVLMs show that current LVLMs exhibit weak fundamental understanding and reasoning capabilities on visual graphs, especially in capturing relational information and dealing with structurally complex visual graphs.

Motivated by the above observations, we further introduce \textsc{MCDGraph}, a novel structure-aware fine-tuning framework to enhance structure learning capabilities of LVLMs through three self-supervised learning tasks: 1) masked graph infilling, 2) contrastive graph discrimination, and 3) graph description.
Experiments show that \textsc{MCDGraph} significantly improves LVLMs' graph understanding and reasoning abilities, especially on edge-related tasks and graph-related downstream tasks. 
Further analysis demonstrates that our method also enhances LVLMs' robustness and generalization to visual graphs with complex structure and unseen styles.
The contributions of this work can be summarized as follows:

\begin{itemize}[itemsep=2pt,topsep=0pt,parsep=0pt,leftmargin=11pt]
    \item We introduce \textsc{VGCure}, a comprehensive benchmark to systematically evaluate LVLMs' fundamental understanding and reasoning abilities on visual graphs.
    \item Through extensive experiments on 14 LVLMs, we reveal LVLMs' limitations in basic graph understanding and reasoning, especially for tasks concerning relational or structural information. 
    \item  We propose a self-supervised framework to enhance LVLMs' ability to capture structural information in visual graphs. Experiments validate its effectiveness on both fundamental and downstream graph learning tasks.
\end{itemize}

\begin{table*}[ht]
   \centering
    \resizebox{\linewidth}{!}{
\begin{tabular}{p{1.5cm}ll}
\toprule
\multicolumn{1}{c}{\textbf{Task}} & \multicolumn{1}{c}{\textbf{QA sample}}                 & \multicolumn{1}{c}{\textbf{FC sample (Label)}}                                      \\
\midrule
\multicolumn{1}{c}{\multirow{2}{*}{\textbf{NNu}}}              & \multirow{2}{*}{\begin{tabular}[c]{@{}l@{}}Q: How many nodes are there in this graph?\\ A: 12\end{tabular}}                                                    & There are 12 nodes in this graph. (True)                                                                              \\ &      & There are 17 nodes in this graph. (False)   \\
\midrule
\multicolumn{1}{c}{\multirow{2}{*}{\textbf{EN}}}               & \multirow{2}{*}{\begin{tabular}[c]{@{}l@{}}Q: How many edges are there in this graph?\\ A: 15\end{tabular}}                                                    & There are 15 edges in this graph. (True)     \\   & & There are 17 edges in this graph. (True)     \\
\midrule
\multicolumn{1}{c}{\multirow{2}{*}{\textbf{DC}}}   & \multirow{2}{*}{-}    & This graph is a directed graph. (True)    \\&    & This graph is an undirected graph. (True)  \\
\midrule
\multicolumn{1}{c}{\multirow{2}{*}{\textbf{DQ}}}      & \multirow{2}{*}{\begin{tabular}[c]{@{}l@{}}Q: What is the degree of E9 in this graph?\\ A: 1\end{tabular}}                                                     & The degree of E9 in this graph is 1. (True)                                                                           \\  &    & The degree of E9 in this graph is 2. (False)    \\
\midrule
\multicolumn{1}{c}{\multirow{2}{*}{\textbf{NQ}}}              & \multirow{2}{*}{\begin{tabular}[c]{@{}l@{}}Q: Which nodes are out-neighbors of E6 in this graph?\\ A: {[}E3{]}\end{tabular}}                                   & E3 is a out-neighbors of E6 in this graph. (True)                                                                     \\ &   & E7 is a out-neighbors of E6 in this graph. (False)      \\  
\midrule
\multicolumn{1}{c}{\multirow{2}{*}{\textbf{NN}}}               & \multirow{2}{*}{\begin{tabular}[c]{@{}l@{}}Q: Which entities are R6 of the entity that is R5 of E3?\\ A: {[}E4, E7{]}\end{tabular}}                            & E4 is R6 of the entity that is R5 of E3. (True)                                                                       \\  &        & E1 is R6 of the entity that is R5 of E3. (False)    \\
\midrule
\multicolumn{1}{c}{\multirow{2}{*}{\textbf{CR}}}               & \multirow{2}{*}{\begin{tabular}[c]{@{}l@{}}Q: Which entities are R4 of E9 as well as R8 of E1?\\ A: {[}E2{]}\end{tabular}}                                     & E2 is R4 of E9 as well as R8 of E1. (True)                                                                            \\ &      & E1 is R4 of E9 as well as R8 of E1. (False)       \\
\midrule
\multicolumn{1}{c}{\multirow{2}{*}{\textbf{CN}}}               & {\multirow{2}{*}{-}}                                                                                                                         & E8 and E1 share a common out-neighbor, i.e., common head entity. (True)    \\  & \multicolumn{1}{c}{}      & E10 and E12 share a common out-neighbor, i.e., common head entity. (False)  \\
\midrule
\multicolumn{1}{c}{\multirow{2}{*}{\textbf{RA}}}               & \multirow{2}{*}{\begin{tabular}[c]{@{}l@{}}Q: Which entities are connected to E3 via the same relation from E3 to E1?\\ A: {[}E2{]}\end{tabular}}           & E2 is connected to E3 via the same relation from E3 to E1. (True)          \\ &      & E6 is connected to E3 via the same relation from E3 to E1. (False)       \\
\midrule
\multicolumn{1}{c}{\multirow{2}{*}{\textbf{SRN}}}              & \multirow{2}{*}{\begin{tabular}[c]{@{}l@{}}Q: Which entities are both R2 of E10?\\ A: {[}E4, E12{]}\end{tabular}}                                              & E4 and E12 are both R2 of E10. (True)                                                                                 \\   &      & E4 and E3 are both R2 of E10. (False)         \\
\midrule
\multicolumn{1}{c}{\multirow{2}{*}{\textbf{NR}}}               & \multirow{2}{*}{\begin{tabular}[c]{@{}l@{}}Q: What is the relation from the entity that is R5 of E3 to E2? \\ A: {[}R8{]}\end{tabular}}                        & E2 is R8 of the entity that is R5 of E3. (True)                                                                       \\ &    & E2 is R4 of the entity that is R5 of E3. (False)    \\
\midrule
\multicolumn{1}{c}{\multirow{2}{*}{\textbf{CP}}}               & \multirow{2}{*}{\begin{tabular}[c]{@{}l@{}}Q: Is there a path from E3 to E4?\\ A: Yes. The paths are {[}{[}E3, E1, E2, E4{]}, {[}E3, E1, E4{]}{]}\end{tabular}} & There are 2 paths from E3 to E4. (True)                                                                               \\   &      & There are 3 paths from E3 to E4. (False)   \\
\bottomrule
\end{tabular}}
 \vspace{-0.2cm}
\caption{Examples for each task in \textsc{VGCure}. These samples all correspond to the graph shown in Fig.\ref{fig:example_graphs}(d).}
\label{tab:examples}
\end{table*}

\section{The \textsc{VGCure} Benchmark}

To evaluate LVLMs' fundamental graph understanding and reasoning capabilities, we introduce \textsc{VGCure}, a large-scale multimodal graph benchmark with 22 challenging tasks.
\textsc{VGCure} features 10 graph types, both synthetic and real-world, to assess LVLMs' performance on diverse graphs. The graphs are anonymized to minimize the impact of pre-existing LLM knowledge on core reasoning abilities, promoting \textit{knowledge-free reasoning} \citep{hu2024large}.
Tab.\ref{tab:introduction} compares \textsc{VGCure} with three recent benchmarks. It is evident that \textsc{VGCure} excels in fundamental graph understanding and reasoning capabilities. 
Furthermore, \textsc{VGCure} offers a comprehensive evaluation through more varied graph types, tasks, and test samples.

\subsection{Graph Structure Generation}
We begin by collecting a wide variety of graph structures for generating visual graphs and challenging tasks. 
Following \citet{fatemi2024test}, we first employ \textit{NetworkX} \citep{hagberg2008exploring} to generate a diverse set of random synthetic structures, including Erdős-Rényi (ER) graphs \citep{erdds1959random}, scale-free networks (SFN) \citep{barabasi1999emergence}, Barabási–Albert (BA) model \citep{albert2002statistical}, stochastic block model (SBM) \citep{holland1983stochastic} and star graphs. 
In addition, we extract anonymized structures from real-world graphs in GraphArena \citep{tang2024grapharena}, including DBLP, Social Network, DBpedia, Openflights and PubChemQC. 
All the entity and relation names in each graph are replaced with generic names to eliminate the impact of the model's internal knowledge on reasoning. 
After initializing the graph structure, we use the \textit{Graphviz} \citep{ellson2002graphviz} to generate concise \textit{directed} and \textit{undirected} visual graphs.

\begin{table*}[t]
   \centering
    \resizebox{\linewidth}{!}{
    \renewcommand\arraystretch{1.2}
\begin{tabular}{lcccccccccccccccccc}
\toprule
\multicolumn{1}{l}{\multirow{3.6}{*}{\textbf{Models}}}   & \multicolumn{5}{c}{\textbf{Understanding}}   & \multicolumn{12}{c}{\textbf{Reasoning}}                                                  \\
\cmidrule(lr){2-6} \cmidrule(lr){7-18}
\multicolumn{1}{c}{}                   & \textbf{NNu}   & \textbf{EN}    & \textbf{DQ}    & \multicolumn{2}{c}{\textbf{NQ}}                    & \multicolumn{2}{c}{\textbf{NN}}  & \multicolumn{2}{c}{\textbf{CR}}  & \multicolumn{2}{c}{\textbf{RA}}  & \multicolumn{2}{c}{\textbf{SRN}} & \multicolumn{2}{c}{\textbf{NR}}  & \multicolumn{2}{c}{\textbf{CP}}   \\
\cmidrule(lr){2-2} \cmidrule(lr){3-3} \cmidrule(lr){4-4} \cmidrule(lr){5-6} \cmidrule(lr){7-8} \cmidrule(lr){9-10} \cmidrule(lr){11-12} \cmidrule(lr){13-14} \cmidrule(lr){15-16} \cmidrule(lr){17-18}
\multicolumn{1}{c}{}            & \textbf{Acc}   & \textbf{Acc}   & \textbf{Acc}   & \textbf{F1}    & \textbf{Hits@1}                      & \textbf{F1}    & \textbf{Hits@1} & \textbf{F1}    & \textbf{Hits@1} & \textbf{F1}    & \textbf{Hits@1} & \textbf{F1}    & \textbf{Hits@1} & \textbf{F1}    & \textbf{Hits@1} & \textbf{EM\_F1} & \textbf{Label\_Acc} \\
\midrule
\textbf{SPHINX}              & 21.03          & 11.38          & 15.45          & 12.84          & 29.45                           & \enspace6.05           & \enspace9.81            & \enspace7.28           & 16.44           & 14.62          & 21.69           & 11.62          & 26.93           & \enspace1.54           & \enspace4.99            & \enspace0.68            & \textbf{95.59}              \\
\textbf{Monkey}             & 40.09          & \enspace9.22           & 17.81          & \enspace9.90           & 21.88                               & \enspace8.90           & 12.96           & \enspace2.85           & \enspace4.62            & \enspace3.74           & \enspace4.13            & \enspace9.79           & 21.61           & \enspace3.06           & \enspace7.98            & \enspace0.46            & \enspace5.11                     \\
\textbf{MiniGPT-v2}               & 11.80          & 10.61          & 18.03          & 16.40          & 27.08                           & \enspace8.92           & 18.02           & \enspace2.01           & \enspace2.65            & \enspace7.50           & 14.87           & \enspace9.85           & 27.47           & \enspace5.86           & 13.87           & \enspace5.26            & \textbf{95.59}         \\
\textbf{mPLUG-Owl3}              & 28.38          & \enspace8.84           & \enspace6.86           & 20.32          & 51.16                              & \enspace8.68           & 11.39           & \enspace5.42           & 11.81           & 11.07          & 14.81           & \enspace6.48           & 18.06           & \enspace2.03           & \enspace0.33            & 10.87           & 74.97             \\
\textbf{LLaVA1.5-7B}                  & 14.53          & \enspace7.56           & 30.14          & 11.43          & 21.25                            & \enspace1.80           & \enspace0.21            & \enspace2.22           & \enspace0.13            & \enspace6.36           & \enspace9.37            & \enspace5.38           & \enspace7.73            & \enspace0.84           & \enspace2.23            & \enspace2.63            & 78.64             \\
\textbf{LLaVA-NeXT}              & 47.89          & \enspace9.59           & 19.52          & 23.27          & 44.11                               & 14.77          & 21.74           & \enspace6.60           & 12.44           & 10.14          & 15.29           & 12.68          & 27.88           & \enspace9.88           & \enspace6.46            & \enspace8.27            & 95.56            \\
\textbf{LLaVA-OV}         & 23.47          & \enspace3.08           & 23.80          & 15.00          & 39.12                          & 10.74          & 19.87           & \enspace7.50           & 16.61           & \enspace9.85           & 19.30           & 10.21          & 30.99           & \enspace1.04           & \enspace1.36            & \enspace9.69            & 94.83                 \\
\textbf{LLaVA1.5-13B}           & 17.08          & \enspace7.62           & 26.33          & 15.68          & 32.66                          & \enspace6.08           & \enspace7.83            & \enspace4.31           & 13.23           & \enspace6.52           & 10.80           & \enspace7.48           & 12.86           & \enspace4.21           & \enspace6.21            & \enspace5.91            & 82.41                     \\
\textbf{InternLM-XC2.5}              & 60.20          & 10.53          & 41.12          & 18.90          & 55.33                   & 14.64          & 26.14           & 19.19          & 45.60           & 11.50          & 19.06           & 14.12          & 46.53           & \enspace3.09           & \enspace5.49            & 28.82  & 95.53             \\
\textbf{Llama3.2}               & 77.31          & \enspace9.39           & 35.56          & 18.79          & 42.34               & \textbf{18.93} & 29.43           & 21.90          & 55.00           & \textbf{15.80} & 27.51           & 20.79 & 61.30           & \textbf{14.11} & \textbf{24.05}  & 20.80           & 94.70            \\
\textbf{Qwen-VL}          & 42.45          & \enspace9.66           & 20.56          & 10.95          & 21.25                                   & 11.44          & 15.51           & \enspace7.93           & 18.30           & 12.48          & 17.36           & 11.11          & 22.79           & \enspace5.43           & \enspace4.01            & \enspace0.00            & \enspace4.48                   \\
\textbf{Qwen2-VL}             & \textbf{97.80} & \textbf{16.38} & 48.09          & 16.18          & 38.12                     & 16.52          & 28.34           & 21.02          & 48.57           & 14.19          & \textbf{27.96}  & 19.48          & 56.42           & 12.73          & 25.07           & 12.90           & 38.06               \\
\textbf{InternVL2}               & 77.45          & \enspace9.78           & 50.75 & 25.01 & 68.58                  & 18.30          & \textbf{30.82}  & \textbf{24.87} & \textbf{59.31}  & 10.83          & 17.12           & 20.72          & \textbf{59.99}  & 10.58          & 18.97           & 14.53           & 43.97                \\
\midrule
\textbf{GPT-4o-mini*}     & 89.20          & 15.40          & \textbf{56.40} & \textbf{30.81} & \textbf{77.40}  & 17.01          & 29.98 & 22.33          & 53.15          & 15.47          & 24.81          & \textbf{22.25} & 59.48          & \enspace8.62           & 13.48          & \textbf{42.55} & 92.40       \\

\bottomrule 
\end{tabular}}
 \vspace{-0.2cm}
        \caption{Model performance on QA samples across various tasks, where \textit{EM\_F1} is the macro F1 score calculated based on the exact match between the predicted path and the ground truth path, \textit{Label\_Acc} measures the accuracy of the model's prediction on whether a path exists or not. The best results are \textbf{bolded}.}
    \label{tab:QA_results}
\end{table*}

\subsection{Tasks Design}
To thoroughly assess the abilities of LVLMs in fundamental graph structure understanding and reasoning, the proposed \textsc{VGCure} encompasses the following categories of tasks. Tab.\ref{tab:examples} presents examples for each task. 

\smallskip

\noindent \textbf{Graph Understanding:} The graph understanding tasks involve analyzing and extracting structural, relational, and property-based information from the visual graph, which aims to gain insights into the composition and topology of the graph, including its nodes, edges, connectivity, and the relationships among its components.
\begin{itemize}[itemsep=2pt,topsep=1pt,parsep=1pt,leftmargin=11pt]
    \item \textbf{Node Number Query (NNu)}: Calculate the total number of nodes in the graph.
    \item \textbf{Edge Number Query (EN)}: Determine the total number of edges in the graph. 
    \item \textbf{Directedness Check (DC)}: Verify whether the graph is undirected or directed.
    \item \textbf{Degree Query (DQ)}: Calculate the degree of the specified node.
    \item \textbf{Neighbor Query (NQ)}: Identify the nodes that are directly connected to the given node.
\end{itemize}

\smallskip

\noindent \textbf{Graph Reasoning:} The reasoning tasks focus on exploring the \textit{knowledge-free reasoning} ability of LVLMs on visual graphs. 
To differentiate from graph understanding tasks, we designed a series of 2-hop reasoning tasks.
\begin{itemize}[itemsep=2pt,topsep=1pt,parsep=1pt,leftmargin=11pt]
    \item \textbf{Nested Node Query (NN)}: Identify the entities linked to the given entity through a composite chain involving the given relations.
    \item \textbf{Conjunctive Relation Query (CR)}: Retrieve the entities satisfying both two independent relationship constraints with two distinct entities.
    \item \textbf{Common Neighbor Check (CN)}: Determine whether two entities share at least one common neighbor in a 2-hop relational path.
    \item \textbf{Relation Analogy Query (RA)}: Find the entities linked to a target entity via a relation identical to that links a given entity pair.
    \item \textbf{Shared Relation Neighbor Query (SRN)}: Identify the set of entities that are connected to the given entity through the given relation.
    \item \textbf{Nested Relation Query (NR)}: Identify the relation between a target entity and an intermediate entity obtained by traversing a specific relation path from a given entity.
    \item \textbf{Connective Path Query (CP)}: Determine the existence of \textit{directed} paths or \textit{shortest undirected} paths between two given entities, and retrieve all possible paths if they exist.
\end{itemize}

\begin{table*}[t]
   \centering
    \resizebox{\linewidth}{!}{
    \renewcommand\arraystretch{1.2}
\begin{tabular}{lcccccccccccccccccccccccc}
\toprule
\multirow{3.6}{*}{\textbf{Models}}                                                                                                 & \multicolumn{10}{c}{\textbf{Understanding}}      & \multicolumn{14}{c}{\textbf{Reasoning}}                                            \\
\cmidrule(lr){2-11} \cmidrule(lr){12-25}
& \multicolumn{2}{c}{\textbf{NNu}} & \multicolumn{2}{c}{\textbf{EN}} & \multicolumn{2}{c}{\textbf{DC}} & \multicolumn{2}{c}{\textbf{DQ}} & \multicolumn{2}{c}{\textbf{NQ}}
                                      & \multicolumn{2}{c}{\textbf{NN}} & \multicolumn{2}{c}{\textbf{CR}} & \multicolumn{2}{c}{\textbf{CN}} & \multicolumn{2}{c}{\textbf{RA}} & \multicolumn{2}{c}{\textbf{SRN}} & \multicolumn{2}{c}{\textbf{NR}} & \multicolumn{2}{c}{\textbf{CP}}                              \\
    \cmidrule(lr){2-3} \cmidrule(lr){4-5} \cmidrule(lr){6-7} \cmidrule(lr){8-9} \cmidrule(lr){10-11} \cmidrule(lr){12-13} \cmidrule(lr){14-15} \cmidrule(lr){16-17} \cmidrule(lr){18-19} \cmidrule(lr){20-21} \cmidrule(lr){22-23} \cmidrule(lr){24-25} 
                                      & \textbf{F1}    & \textbf{Acc}   & \textbf{F1}    & \textbf{Acc}   & \textbf{F1}    & \textbf{Acc}   & \textbf{F1}    & \textbf{Acc}   & \textbf{F1}     & \textbf{Acc}   & \textbf{F1}    & \textbf{Acc}   & \textbf{F1}    & \textbf{Acc}   & \textbf{F1}     & \textbf{Acc}   & \textbf{F1}    & \textbf{Acc}   & \textbf{F1}    & \textbf{Acc}   & \textbf{F1}    & \textbf{Acc}   & \textbf{F1}    & \textbf{Acc}              \\

\midrule
\textbf{SPHINX}             & 33.33           & 50.00          & 33.33          & 50.00          & 33.33          & 50.00          & 33.33          & 50.00          & 33.33          & 50.00                 & 33.33          & 50.00          & 33.33          & 50.00          & 33.84          & 51.16          & 33.33          & 50.00          & 33.33           & 50.00          & 33.33          & 50.00          & 33.33          & 50.00              \\
\textbf{Monkey}           & 40.31           & 53.23          & 51.38          & 55.58          & 64.25          & 65.24          & 50.01          & 55.62          & 44.04          & 53.29                   & 50.54          & 50.76          & 37.37          & 51.08          & 48.85 & 49.71          & 38.32          & 50.13          & 33.52           & 49.78          & 40.29          & 48.05          & 51.44          & \textbf{58.87}                \\
\textbf{MiniGPT-v2}        & 34.37           & 50.38          & 34.28          & 50.39          & 33.92          & 50.18          & 33.69          & 49.91          & 44.29          & 50.99                    & 49.68          & 50.03          & 47.94          & 53.19          & 45.66          & 51.80 & 36.71          & 49.48          & 43.08           & 43.22          & 36.68          & 50.07          & 51.68          & 52.39                      \\
\textbf{mPLUG-Owl3}         & 37.53           & 50.80          & 31.16          & 38.59          & 73.04          & 74.76          & 39.84          & 46.92          & 33.33          & 50.00                & 34.75          & 49.86          & 46.11          & 52.04          & 33.96          & 51.20          & 44.80          & 48.54          & 47.35           & 53.10          & 36.65          & 49.30          & 37.34          & 50.27                      \\
\textbf{LLaVA1.5-7B}             & 33.33           & 50.00          & 33.33          & 50.00          & 33.33          & 50.00          & 33.33          & 50.00          & 33.33          & 50.00             & 33.33          & 50.00          & 33.33          & 50.00          & 33.84          & 51.16          & 33.33          & 50.00          & 33.33           & 50.00          & 33.33          & 50.00          & 33.33          & 50.00                      \\
\textbf{LLaVA-NeXT}         & 33.33           & 50.00          & 33.33          & 50.00          & 33.33          & 50.00          & 33.33          & 50.00          & 33.33          & 50.00             & 33.33          & 50.00          & 33.33          & 50.00          & 33.84          & 51.16          & 33.33          & 50.00          & 33.33           & 50.00          & 33.33          & 50.00          & 33.33          & 50.00                   \\
\textbf{LLaVA-OV}       & 39.13           & 52.69          & 33.79          & 49.68          & 65.02          & 68.49          & 36.87          & 51.49          & 37.12          & 51.70           & 46.83          & 53.37          & 39.66          & 52.41          & 34.09          & 51.23          & 39.64          & 48.89          & 44.12           & 52.19          & 48.19          & 52.19          & 33.35          & 49.05                      \\
\textbf{LLaVA1.5-13B}         & 60.31           & 63.26          & 50.78          & 57.09          & 87.74          & 87.92          & 33.33          & 50.00          & 36.39          & 49.31                 & 47.65          & 50.31          & 35.93          & 50.94          & 34.68          & 49.31          & 33.33          & 50.00          & 33.47           & 50.03          & 35.18          & 43.68          & 44.55          & 52.91                    \\
\textbf{InternLM-XC2.5}      & 37.19           & 51.79          & 40.00          & 51.05          & 65.76          & 66.72          & 43.07          & 49.16          & 64.46          & 65.40         & 50.81          & 54.05          & 58.72          & 61.66          & 39.34          & 50.05          & 35.54          & 50.47          & 49.36           & 51.98          & 49.12          & 53.08          & 52.12          & 52.76              \\
\textbf{Llama3.2}   & 87.66           & \textbf{87.66}          & 47.26          & 49.67          & 43.27          & 51.83          & 42.23          & 51.39          & 65.87          & 66.03    & 53.47 & 56.40          & 62.79          & 63.74          & 39.00          & 50.30          & \textbf{59.89} & \textbf{61.20} & 48.82           & 49.38          & \textbf{58.59}          & \textbf{60.36} & 45.65          & 49.32           \\
\textbf{Qwen-VL}       & 32.30           & 47.72          & 32.27          & 47.56          & 9.50           & 10.49          & 31.04          & 44.54          & 36.69          & 49.97                  & 27.07          & 32.57          & 15.55          & 17.99          & 31.68          & 45.30          & 27.83          & 29.13          & 32.71           & 33.70          & 34.72          & 41.94          & 36.37          & 42.63                     \\
\textbf{Qwen2-VL}        & \textbf{76.50}  & 77.67 & \textbf{68.26} & \textbf{68.27} & \textbf{94.92} & \textbf{94.94} & \textbf{64.32} & \textbf{67.40} & 67.34          & 68.76         & 44.17          & 53.77          & 74.28          & 75.24          & 38.52          & 51.10          & 35.28          & 50.55          & \textbf{57.07}  & \textbf{59.18} & 42.08          & 52.95          & 42.25          & 53.38        
 \\
 \textbf{InternVL2}            & 68.63           & 71.18          & 36.82          & 50.23          & 93.17          & 93.18          & 63.28          & 63.84          & \textbf{72.81} & \textbf{73.27}         & \textbf{62.46}          & \textbf{63.62} & \textbf{75.37} & \textbf{76.18} & 47.12          & 50.40          & 33.70          & 50.14          & 56.98           & 57.42          & 55.94 & 58.42          & \textbf{54.71} & 54.71                    \\
\midrule
\textbf{GPT-4o-mini*}   & 64.49          & 67.00          & 39.85          & 52.30          & 90.52          & 90.60          & 42.03          & 52.90          & 48.77          & 56.80        & 50.77          & 51.51          & 62.17          & 62.82          & \textbf{53.03} & \textbf{55.05} & 36.50          & 50.78          & 54.87          & 54.89          & 44.37          & 51.11          & 51.39          & 54.50    
\\
\bottomrule 
\end{tabular}}
 \vspace{-0.2cm}
        \caption{Model performance on FC samples across various tasks. The best results are \textbf{bolded}.}
    \label{tab:FC_results}
\end{table*}

For each task, we construct \textbf{one QA sample} and \textbf{two fact checking (FC) samples} (with labels of True and False, respectively) automatically based on the template, except for CN and DC which have only fact checking samples due to the strong similarity between the two samples. The total number of final samples is 223,646. More details about the \textbf{design logic} and \textbf{statistics} of \textsc{VGCure} can be found in Appendix \ref{app:benchmark} and Tab.\ref{tab:statistics}.

\section{Benchmarking LVLMs on \textsc{VGCure}}
\subsection{Experimental Setup}
We conduct evaluation on 13 open-source LVLMs, including InternLM-XComposer2.5-7B \citep{zhang2024internlm}, InternVL2-8B, Llama3.2-11B-Vision-Instruct, LLaVA1.5-7B \citep{Liu_2024_CVPR}, LLaVA1.5-13B \citep{Liu_2024_CVPR}, LLaVA-NeXT-7B \citep{li2024llava}, LLaVA-OneVision-7B \citep{li2024llavaonevision}, MiniGPT-v2 \citep{chen2023minigpt}, Monkey \citep{Li_2024_CVPR}, mPLUG-Owl3-7B \citep{ye2024mplug}, Qwen-VL \citep{bai2023qwen}, Qwen2-VL-7B-Instruct \citep{wang2024qwen2} and SPHINX \citep{lin2023sphinx}. 
Meanwhile, we also evaluate the performance of the GPT-4o-mini, which is a strong closed-source LVLM\footnote{We excluded GPT-4o as a baseline due to its high cost.}.
Due to the high cost of GPT-4o-mini, we randomly select 50 graphs from each graph structure for testing. 
For all methods, the \textbf{zero-shot} setting is adopted during evaluation. 
More details can be found in Appendix \ref{app:evaluation_experimental_setup}.

\begin{figure*}[t]
    \centering
    \includegraphics[width=0.95\linewidth]{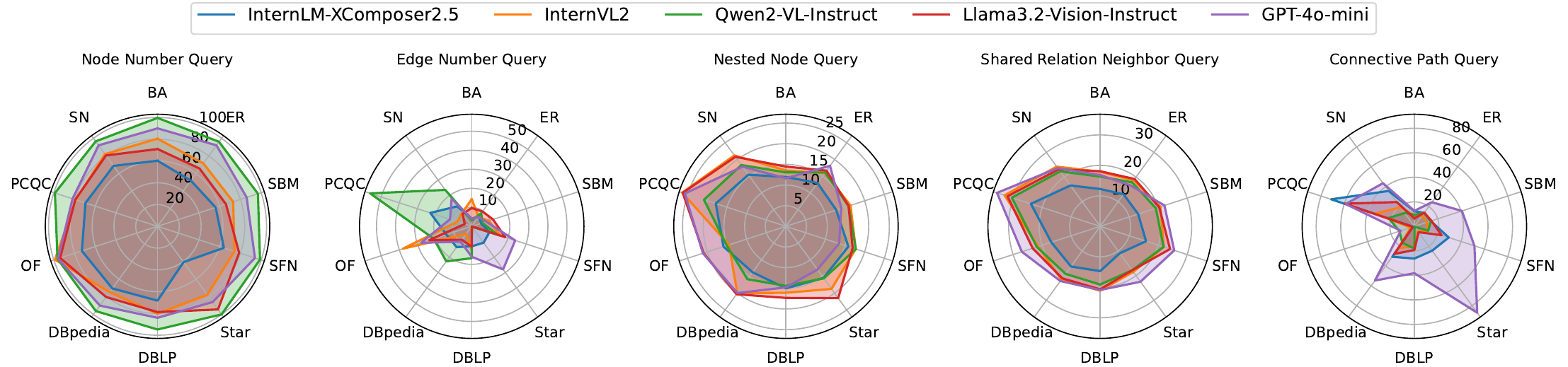}
     \vspace{-0.2cm}
    \caption{Model performance (F1/Acc) on representative QA samples across various graph structures and tasks
    , where OF, PCQC and SN denotes Openflights, PubChemQC and Social network, respectively.}
    \label{fig:graph_results}
\end{figure*}

\begin{figure*}[t]
    \centering
    \subfigure[Number of Edges]{
        \includegraphics[width=0.32\textwidth]{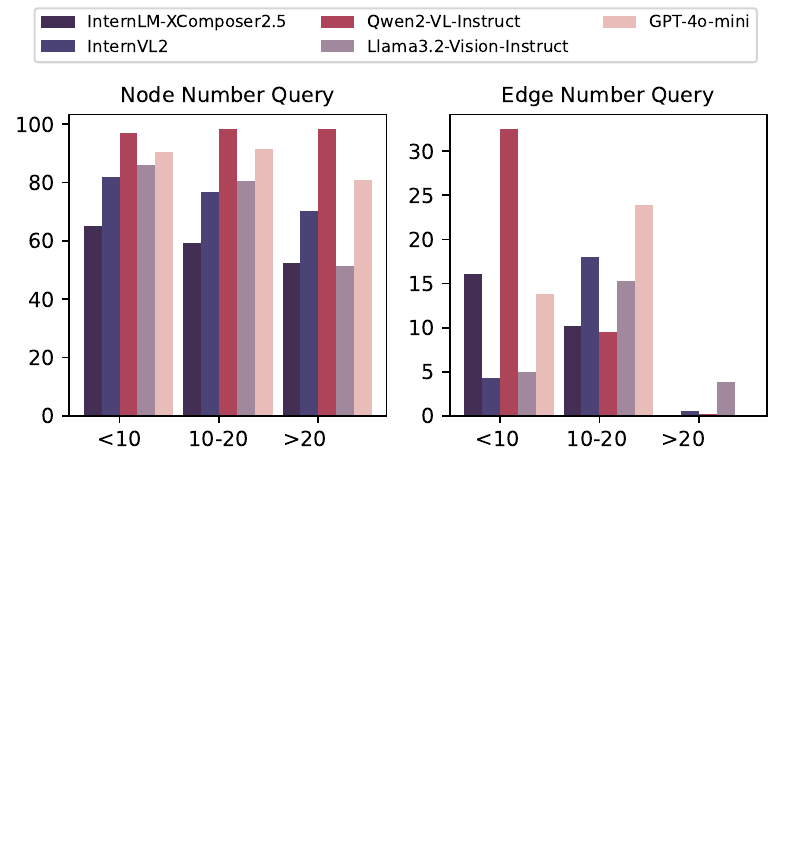}
    }
    \hspace{-0.3cm}
    \subfigure[Number of Nodes]{
        \includegraphics[width=0.32\textwidth]{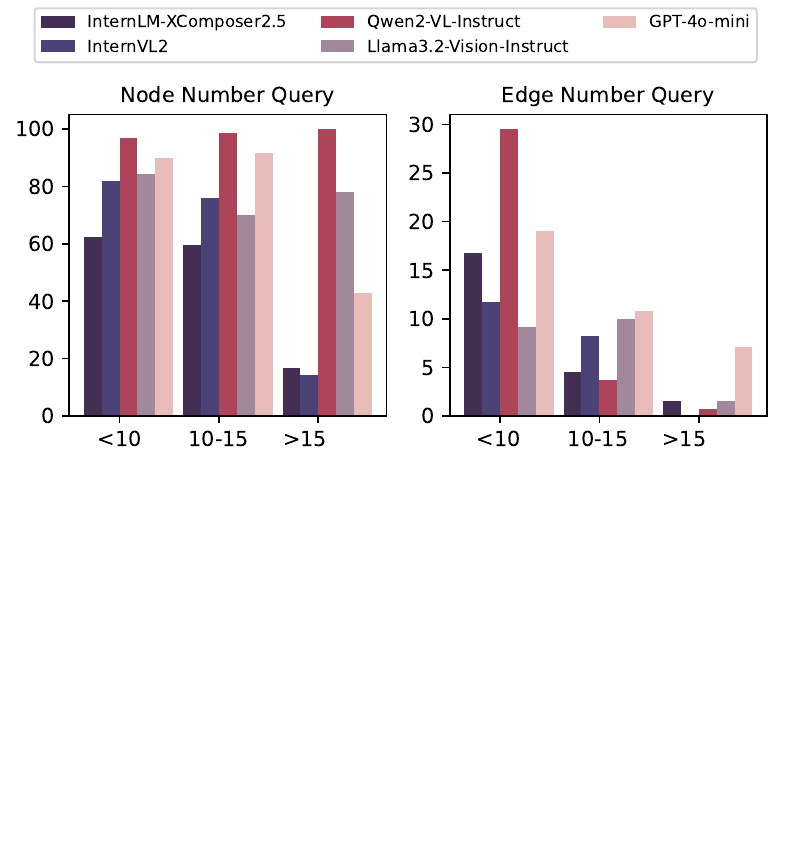}
    }
    \hspace{-0.3cm}
    \subfigure[Average Degree]{
        \includegraphics[width=0.32\textwidth]{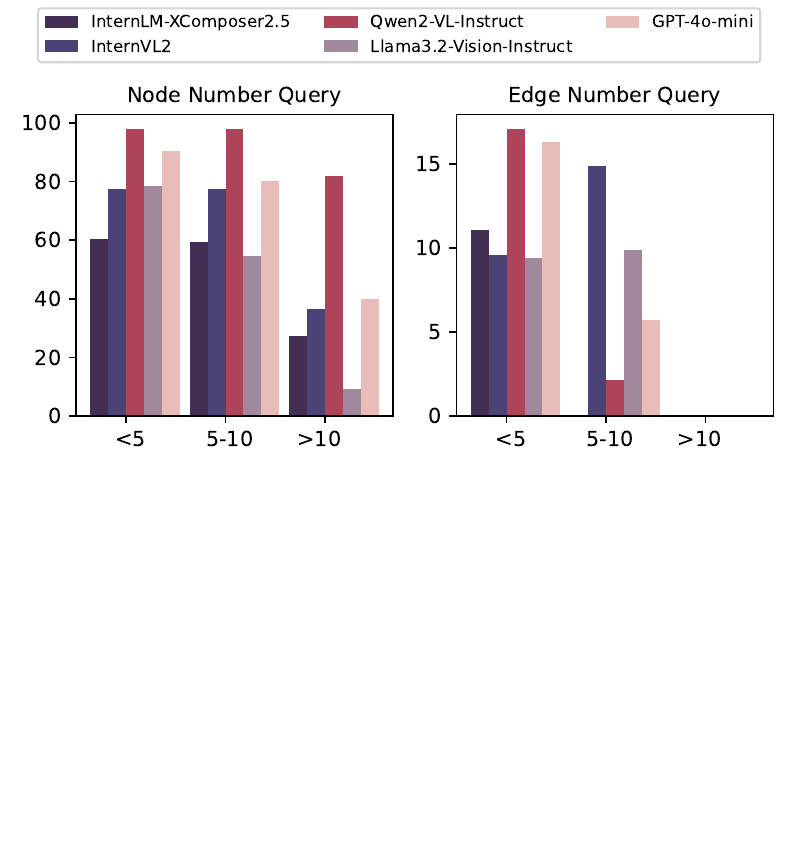}
    }
     \vspace{-0.2cm}
    \caption{Model performance (Acc) comparison on QA samples across various dimensions of complexity.}
    \label{fig:complexity_results}
\end{figure*}

\subsection{Main Result}
\noindent\textbf{Graph Understanding} 
Tab.\ref{tab:QA_results} and Tab.\ref{tab:FC_results} present the evaluation results for question answering (QA) and fact checking (FC), respectively.
We report the averaged results across various graph structures.
In general, most LVLMs struggle to precisely understand the structural and relational information in visual graphs.
In specific, 
(\uppercase\expandafter{\romannumeral1}) Among the graph understanding tasks, \textit{Node Number Query (NNu)} and \textit{Directedness Check (DC)} are the easiest for most LVLMs. This indicates that most LVLMs can accurately capture the number of nodes and directedness information within the visual graph.
(\uppercase\expandafter{\romannumeral2}) All LVLMs struggle with \textit{Edge Number Query (EN)} and \textit{Neighbor Query (NQ)} tasks, with the highest accuracy of 16.38\% and F1 score of 30.81\% , respectively. 
This indicates that current LVLMs are weak in understanding relational and structural information.
(\uppercase\expandafter{\romannumeral3}) Even with a similar number of parameters, the graph understanding abilities of open-source LVLMs vary significantly, with Qwen2-VL and InternVL2 showing better performance in both QA and FC samples.
(\uppercase\expandafter{\romannumeral4}) For the same task, LVLMs perform differently on QA and FC samples, likely due to different ways in reasoning and understanding required by each task \citep{thorne-etal-2018-fever}.
(\uppercase\expandafter{\romannumeral5}) The closed-source LVLM, GPT-4o-mini, offers no significant advantages and even underperforms open-source LVLMs, especially on FC tasks.

\noindent\textbf{Graph Reasoning} 
Based on graph reasoning results in Tabs.\ref{tab:QA_results} and \ref{tab:FC_results}, it can be observed that,
(\uppercase\expandafter{\romannumeral1}) Compared to graph understanding, the graph reasoning tasks are more challenging and the overall performance of LVLMs is worse on both QA and FC samples. This may be a knock-on effect due to deficiencies in visual graph understanding.
(\uppercase\expandafter{\romannumeral2}) Among all open-source LVLMs, InternVL2 and Llama3.2-Vision achieve better performance and LLaVA1.5-7B perform the worst on graph reasoning tasks.
(\uppercase\expandafter{\romannumeral3}) All the LVLMs perform poorly on \textit{Nested Relation Query (NR)} for both QA and FC samples, which is similar to the observation in the graph understanding task. 
This suggests that LVLMs are deficient in recognizing edges and understanding structural information within visual graphs.
(\uppercase\expandafter{\romannumeral4}) On the QA samples, the performance of different LVLMs on \textit{Connective Path Query (CP)} varies widely. SPHINX, Monkey and Qwen-VL demonstrate almost no ability to recognize paths between two specific nodes in the visual graph.
(\uppercase\expandafter{\romannumeral5}) Similarly, GPT-4o-mini underperforms in most tasks compared to open-source LVLMs, except for the \textit{Connective Path Query (CP)} task.

\subsection{Impact of Structures}
Inspired by \citet{fatemi2024test}, we explore how graph structure affects LVLMs' ability to understand and reason on visual graphs. 
Fig.\ref{fig:graph_results} compares the performance of the top five LVLMs on QA samples across various structures.
Obviously, the graph structure significantly impacts LVLMs' performance on most tasks.
Notably, all LVLMs perform well on \textit{PCQC}, which has a simpler structure with fewer nodes and edges (averaging 5.45 nodes and 4.76 edges), and weaker on \textit{BA}, which has the highest edge count in \textsc{VGCure} (averaging 21.02 edges).
Furthermore, on different tasks, the performance of LVLMs is affected differently by the graph structure.
\textit{EN} and \textit{CP} show larger performance variations across graph structures, whereas \textit{NNu} and \textit{NN} show smaller differences.
More results are shown in Figs.\ref{fig:graph_results_qa} and \ref{fig:graph_results_fc}, the overall trends are similar to above findings.


\subsection{Impact of Complexity}
We further discuss the impact of graph complexity on the LVLMs' ability to understand and reason over the visual graph by considering the number of edges, number of nodes, and average degree. 
Fig.\ref{fig:complexity_results} shows a performance comparison of five LVLMs on QA samples across representative tasks and complexity levels. 
As complexity increases, LVLMs' performance declines, especially on \textit{EN}, where results vary significantly.
Besides, some LVLMs perform best at intermediate complexity, but struggle with more complex graphs. This reflects a balance between information richness and complexity of visual graphs, whereas higher complexity likely overwhelm the LVLMs' abilities to reason or generalize due to the complex information within large graphs.
In addition, different complexity dimensions affect LVLMs' performance on various tasks differently, such as \textit{NNu}, where LVLMs are more influenced by the number of nodes and average degree than by edges.
More results in Figs.\ref{fig:size_results_qa}-\ref{fig:degree_results_fc} show the similar trend.


\begin{figure}[t]
    \centering
    \subfigure{
        \includegraphics[width=0.20\textwidth]{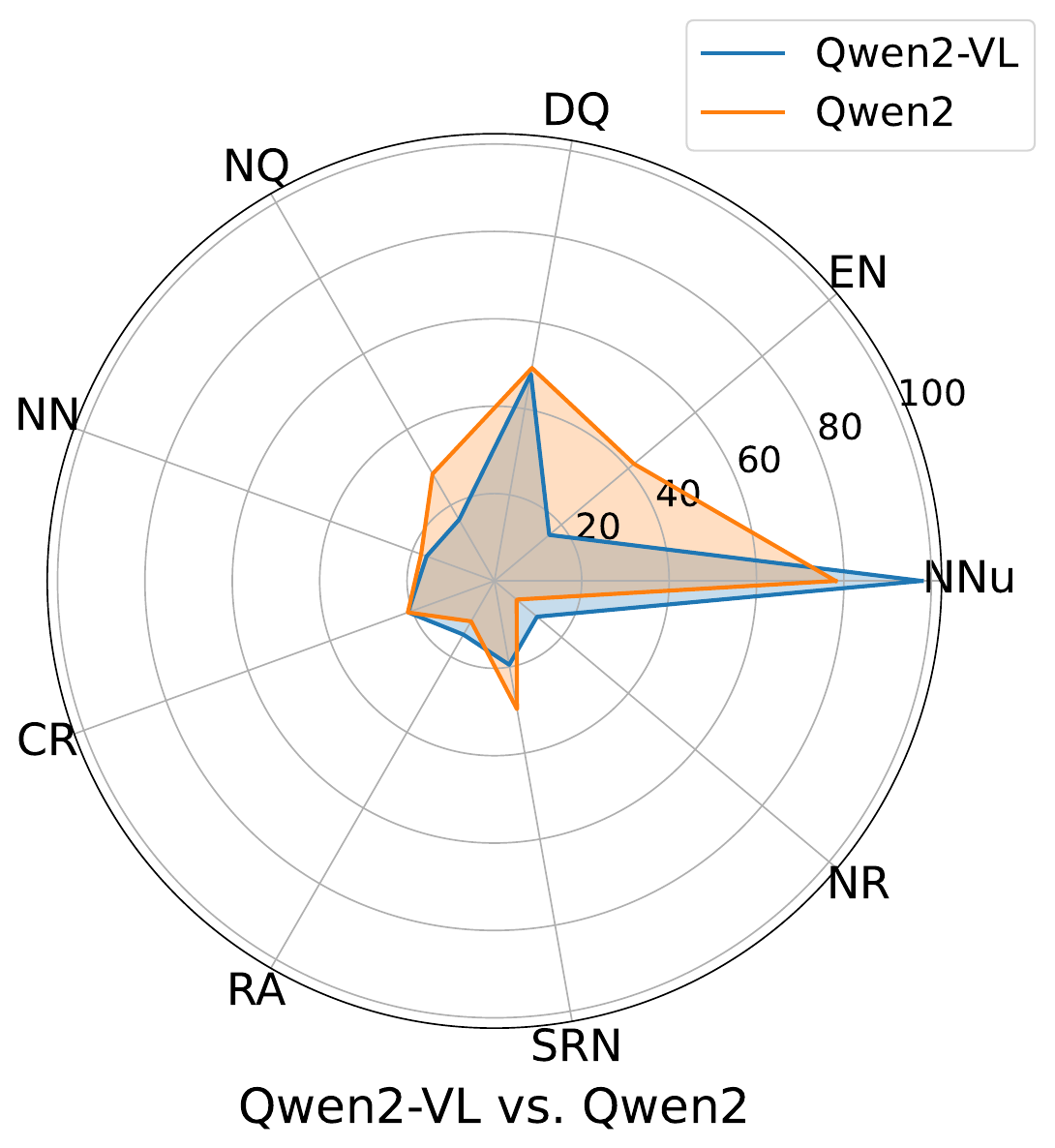}
    }
    \hspace{-0.0cm}
    \subfigure{
        \includegraphics[width=0.20\textwidth]{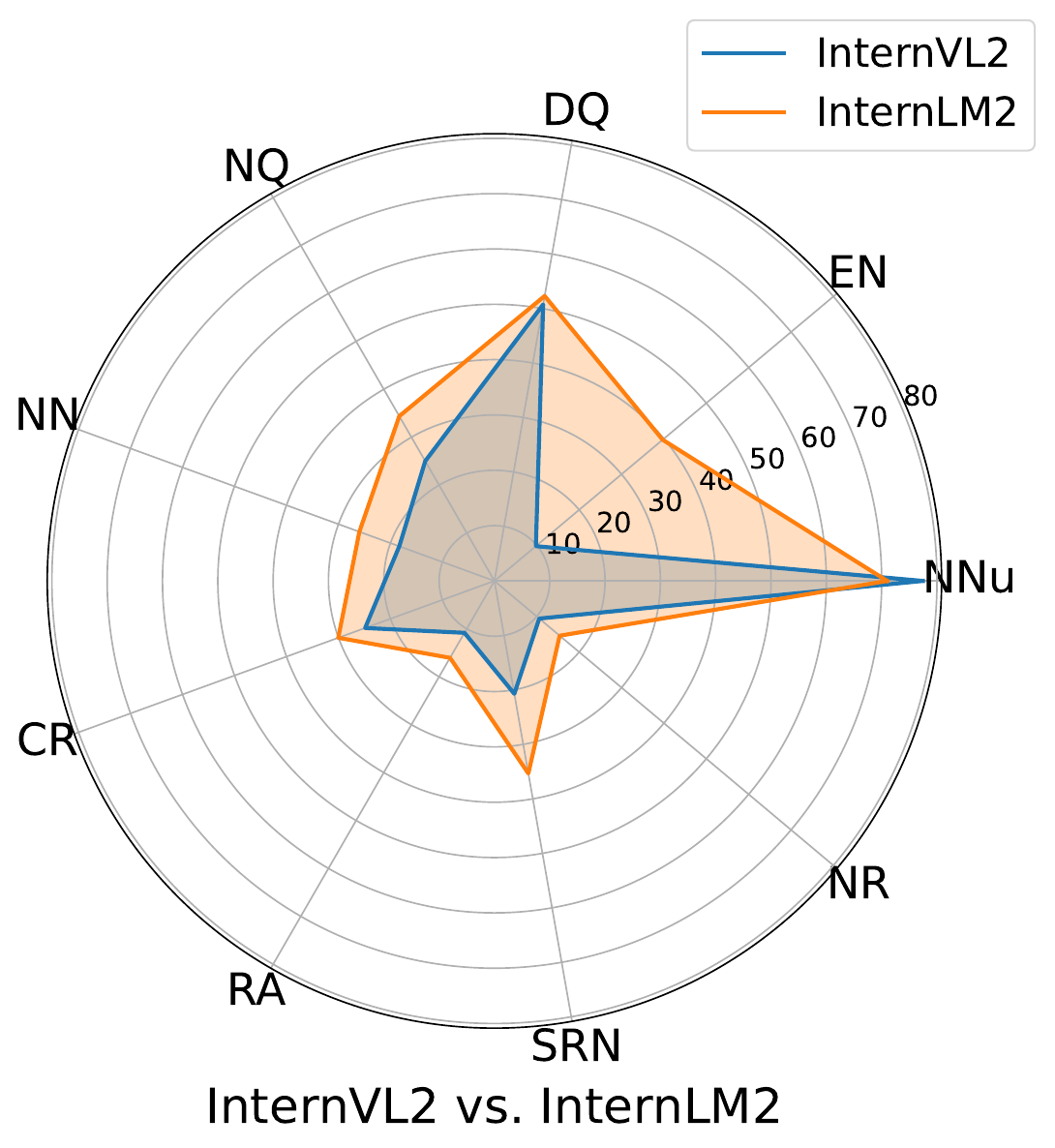}
    }
     \vspace{-0.4cm}
    \caption{Results comparison between LVLM and LLM on QA samples across various tasks.}
    \label{tab:text_results}
\end{figure}

\subsection{Why LVLMs Fail on Fundamental Tasks?}
To explore in depth whether the failure of LVLMs on the fundamental task is due to their weak \textbf{perception} or \textbf{other higher-order} ability, we compare the performance of LVLMs and the corresponding backbone LLMs that take the textual graph as input. 
As Fig.\ref{tab:text_results} shows, despite the disparity in the performance of the individual models, LVLMs perform weaker than the corresponding backbone LLMs on both understanding and reasoning tasks. 
In particular, LVLMs' performance degrades the most on \textit{Edge Number Query (EN)}, which also illustrates the lack of ability of LVLMs to capture structural information in the visual graph.
This indicates that the failure of LVLMs is partly attributed to their \textbf{weaker visual graph perception}. 
In addition, the performance of the backbone LLMs on these tasks remains undesired, with all accuracies below 45\%, this suggests that the backbone LLMs exhibit limited capabilities in understanding and reasoning on graph data. 
More details and results are presented in Appendix \ref{app:text_comparison}.

\begin{table}[t]
    \centering
        \resizebox{\linewidth}{!}{
    \begin{tabular}{lcccccc}
    \toprule
    \textbf{Error Type} & \textbf{R.M.} & \textbf{C.L.} & \textbf{S.H.} & \textbf{E.A.} & \textbf{Ot.G.} & \textbf{F.E.} \\
    \midrule
    Percentage          & 54\%          & 7\%           & 17\%          & 9\%           & 5\%            & 8\%  \\
    \bottomrule
    \end{tabular}}
    \vspace{-0.2cm}
    \caption{Frequency of each error type.}
    \label{tab:error_frequency}
\end{table}

\subsection{Error Analysis}
In addition, by analyzing a batch of results generated by Qwen2-VL and InternVL2 on QA samples in \textsc{VGCure}, we find that their main errors can be grouped into the following six categories:
\begin{itemize}[itemsep=2pt,topsep=1pt,parsep=1pt,leftmargin=11pt]
    \item \textbf{Relation Misunderstanding (R.M.)}: Failure to properly understand the relations between entities, leading to erroneous reasoning.
    \item \textbf{Complexity Limitation (C.L.)}: When faced with intricate or highly complex graph structures, LVLMs struggle to process and understand the information effectively, often resulting in incomplete or erroneous outputs.
    \item \textbf{Structural Hallucination (S.H.)}: Generating or perceiving structures that do not exist, leading to erroneous or misleading results that do not match the actual visual graph.
    \item \textbf{Entity-based Answering (E.A)}: Directly using the entities mentioned in the question as answers, thus ignoring deeper relation understanding or logical reasoning in the visual graph.
    \item \textbf{Off-target Generation (Ot.G.)}: Deviation from the task or misunderstanding of the question, leading to the generation of irrelevant answers.
    \item \textbf{Format Error (F.E.)}: The output of the model is incorrectly formatted or unexpected.
\end{itemize}
Examples of each error are shown in the Tab.\ref{tab:error_type}.

\noindent\textbf{Error Distribution Analysis} We also calculate the overall distribution of each error. The results are shown in Tab.\ref{tab:error_frequency}. It can be observed that, (\uppercase\expandafter{\romannumeral1}) Relation Misunderstanding appears most frequently due to LVLMs' limited capacity to effectively capture structural and relational information in the visual graph.
(\uppercase\expandafter{\romannumeral2}) Structural Hallucination also appears frequently due to the inherent hallucination tendency of LLMs.
(\uppercase\expandafter{\romannumeral3}) Despite demonstrating robust instruction-following capabilities, LVLMs remain prone to errors like Off-target Generation and Format Error.
(\uppercase\expandafter{\romannumeral4}) Complexity Limitation and Entity-based Answering also account for a certain percentage, i.e., LVLMs may be stumped when faced with overly complex visual graphs, or directly use the entities mentioned in the question as answers.

\noindent\textbf{Primary Error Pattern Analysis} We further analyze the primary error pattern of each task. We find that, 
(\uppercase\expandafter{\romannumeral1}) Generally, the primary error occurring in all tasks except the \textit{Node Number Query} is Relation Misunderstanding.
(\uppercase\expandafter{\romannumeral2}) Structural Hallucination occurs more frequently in \textit{Nested Node Query}, \textit{Conjunctive Relation Query,} \textit{Relation Analogy Query}, \textit{Shared Relation Neighbor Query}, and \textit{Nested Relation Query} tasks.
(\uppercase\expandafter{\romannumeral3}) Format Error is usually found in \textit{Connective Path Query} task.

\section{The \textsc{MCDGraph} Framework}
\begin{figure}[t]
    \centering
    \includegraphics[width=\linewidth]{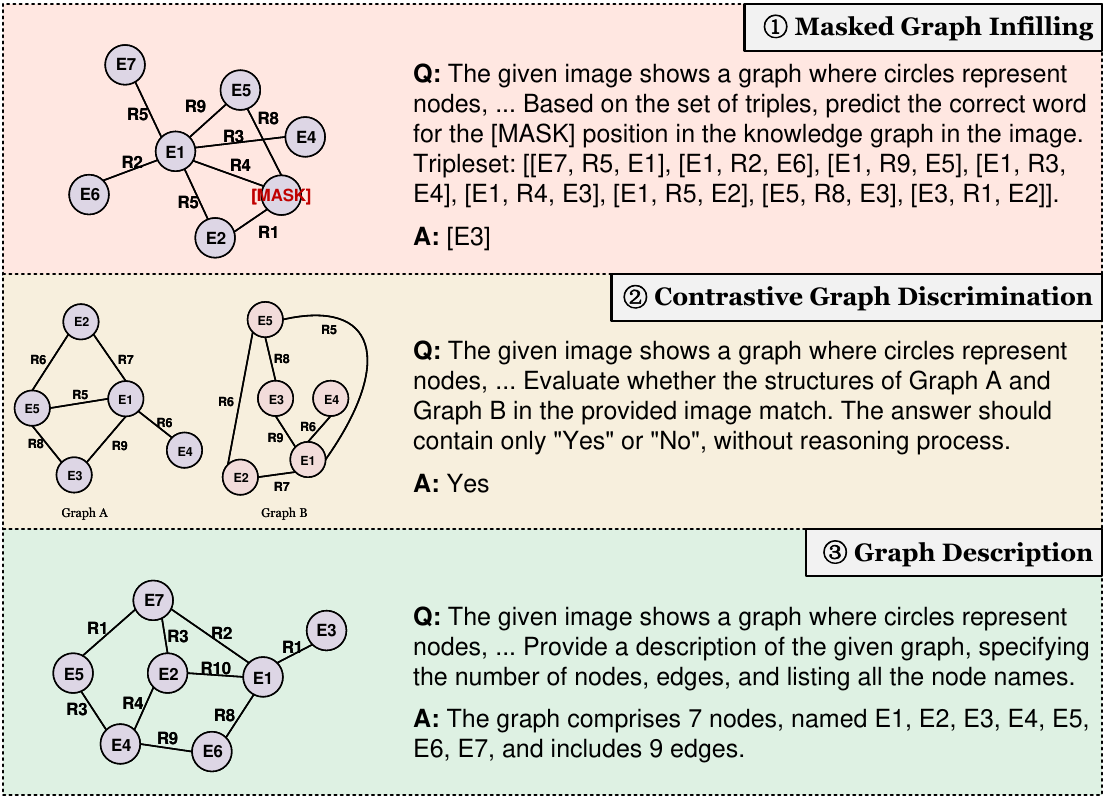}
     \vspace{-0.8cm}
    \caption{Overall illustration of \textsc{MCDGraph}.}
    \label{fig:MCDGraph}
\end{figure}

To enhance the ability of LVLMs to understand and reason on visual graphs, we propose \textsc{MCDGraph}, a \textbf{self-supervised} fine-tuning framework designed to improve LVLMs' ability to capture structural and relational information within visual graphs. 
As illustrated in Fig.\ref{fig:MCDGraph}, \textsc{MCDGraph} comprises three key tasks: \textbf{M}asked Graph Infilling, \textbf{C}ontrastive Graph Discrimination, and Graph \textbf{D}escription.

\begin{table*}[t]

   \centering
    \resizebox{\linewidth}{!}{
\begin{tabular}{l|cccccccccccc}
\toprule
\multirow{2.5}{*}{\textbf{Models}} & \multicolumn{5}{c}{\textbf{Understanding}}                           & \multicolumn{7}{c}{\textbf{Reasoning}}                                                           \\  \cmidrule(lr){2-6}  \cmidrule(lr){7-13}
                                 & \textbf{NNu} & \textbf{EN} & \textbf{DC} & \textbf{DQ} & \textbf{NQ} & \textbf{NN} & \textbf{CR} & \textbf{CN} & \textbf{RA} & \textbf{SRN} & \textbf{NR} & \textbf{CP} \\
                                 \midrule
                                 & \multicolumn{12}{c}{\textbf{QA Samples}}         \\
                                \midrule
\textbf{Qwen2-VL}                & 97.80        & 16.38       & -           & 48.09       & 16.18       & 16.52       & 21.02       & -           & 14.19       & 19.48        & 12.73       & 12.90       \\
\rowcolor{blue!10}\textbf{w \textsc{MCDGraph}}              & 98.34 $\uparrow$       & 25.92 $\uparrow$      & -           & 60.94 $\uparrow$      & 25.44 $\uparrow$      & 13.32       & 26.14 $\uparrow$      & -           & 13.14       & 20.74 $\uparrow$       & 14.44   $\uparrow$    & 11.95       \\
\textbf{InternVL2}               & 77.45        & 9.78        & -           & 50.75       & 25.01       & 18.30       & 24.87       & -           & 10.83       & 20.72        & 10.58       & 14.53       \\
\rowcolor{blue!10}\textbf{w \textsc{MCDGraph}}              & 95.68  $\uparrow$      & 40.45 $\uparrow$      & -           & 54.78  $\uparrow$     & 28.80  $\uparrow$     & 19.43  $\uparrow$     & 28.53  $\uparrow$     & -           & 11.67 $\uparrow$      & 22.34 $\uparrow$       & 16.50 $\uparrow$      & 12.76       \\
\midrule
                                 & \multicolumn{12}{c}{\textbf{FC Samples}}   \\\midrule
\textbf{Qwen2-VL}                & 76.50        & 68.26       & 94.92       & 64.32       & 67.34       & 44.17       & 74.28       & 38.52       & 35.28       & 57.07        & 42.08       & 42.25       \\
\rowcolor{blue!10} \textbf{w \textsc{MCDGraph}}              & 89.58 $\uparrow$       & 65.80       & 95.84 $\uparrow$      & 77.10 $\uparrow$      & 79.75 $\uparrow$      & 60.71 $\uparrow$      & 83.07 $\uparrow$      & 53.90 $\uparrow$      & 53.48 $\uparrow$      & 64.17 $\uparrow$       & 64.12 $\uparrow$      & 60.11  $\uparrow$     \\
\textbf{InternVL2}               & 68.63        & 36.82       & 93.17       & 63.28       & 72.81       & 62.46       & 75.37       & 47.12       & 33.70       & 56.98        & 55.94       & 54.71       \\
\rowcolor{blue!10}\textbf{w \textsc{MCDGraph}}              & 76.55 $\uparrow$       & 71.98 $\uparrow$      & 90.04       & 56.83       & 80.98 $\uparrow$      & 73.14 $\uparrow$      & 80.23  $\uparrow$     & 52.09  $\uparrow$     & 52.81 $\uparrow$      & 59.07 $\uparrow$       & 69.03 $\uparrow$      & 45.82    \\
\bottomrule
\end{tabular}}
\vspace{-0.2cm}
        \caption{Model performance (Acc/F1/EM\_F1 for QA and F1 for FC) on various tasks. $\uparrow$ indicates an improvement compared to the original model. The complete experimental results are shown in Tab.\ref{tab:MCDGraph_results_qa_all} and \ref{tab:MCDGraph_results_fc_all}.}
            \label{tab:MCDGraph_results}
\end{table*}

\subsection{Task1: Masked Graph Infilling}
For this task, we randomly mask either nodes or edges in a visual graph and challenge the model to predict the masked element based on the partially observed graph.
Since anonymized visual graphs contain no semantic information, we also provide the corresponding text triples as input.
\begin{equation}
\setlength\abovedisplayskip{4pt}
\setlength\belowdisplayskip{4pt}
    M = \text{LVLMs}(\hat{G}, I, T),
\end{equation}
where
$\hat{G}$, $I$, $T$ denotes the masked graph, task instruction, and text triples of the original graph, respectively.
This task encourages LVLMs to infer missing structure information, improving their ability to understand graph structure and the relationships between elements.

\subsection{Task2: Contrastive Graph Discrimination}
To further refine the LVLMs' understanding of graph structure, we introduce a contrastive learning task, which helps train the LVLMs to distinguish between two visual graphs that may either represent the \textit{same graph with different layouts} or two \textit{distinct graphs with similar layouts}. 
\begin{equation}
\setlength\abovedisplayskip{4pt}
\setlength\belowdisplayskip{4pt}
    Y = \text{LVLMs}(G_1, G_2, I),
\end{equation}
where the answer $Y\in\{\textit{Yes, No}\}$, $G_1, G_2$ denotes two graphs, and $I$ is the task instruction.
By learning how to perform structural reasoning and graphical isomorphism detection, this task aims to improve LVLMs by recognizing subtle structural differences between two visual graphs.

\subsection{Task3: Graph Description}
Graph Description task requires LVLMs to generate a textual description of a given visual graph,
including the total number of nodes and edges, as well as the names of all the nodes in the graph,
\begin{equation}
\setlength\abovedisplayskip{4pt}
\setlength\belowdisplayskip{4pt}
    D = \text{LVLMs}(G, I),
\end{equation}
where $D$ represents the description, and $G$, $I$ denotes the input graph and task instruction, respectively.
This task ensures that the LVLMs develop a clear understanding of the graph's composition, thereby enhancing their ability to interpret and summarize graph-based information.

\section{Enhancing LVLMs with \textsc{MCDGraph}}
\subsection{Experimental Setup}

We validate the effectiveness of \textsc{MCDGraph} on top two performing 
 LVLMs on \textsc{VGCure}, i.e., Qwen2-VL and InternVL2. 
We collect a new set of anonymized visual graphs \textbf{beyond \textsc{VGCure}} with synthetic structures and automatically create 20k training samples for \textsc{MCDGraph}. 
To prevent catastrophic forgetting, we apply LoRA \citep{hu2022lora} to efficiently enhance the LVLMs' abilities while preserving their original performance.
More details about training samples and implementation are available in Appendix \ref{app:training_experimental_setup}.

\subsection{Results and Analysis}
\noindent \textbf{Main Results}
Tab.\ref{tab:MCDGraph_results} compares the performance of Qwen2-VL and InternVL2 before and after applying \textsc{MCDGraph} on both visual graph understanding and reasoning tasks.
We can observe that 
(\uppercase\expandafter{\romannumeral1}) \textsc{MCDGraph} improves the performance of LVLMs on almost all tasks, demonstrating the effectiveness of the proposed method.
(\uppercase\expandafter{\romannumeral2}) The improvement of LVLMs is particularly impressive on edge-related tasks, i.e., \textit{Edge Number Query (EN)} and \textit{Nested Relation Query (NR)}, that are relatively difficult for LVLMs. This suggests that \textsc{MCDGraph} enhances LVLMs' ability to capture structural information.
(\uppercase\expandafter{\romannumeral3}) Although \textsc{MCDGraph} does not optimize for the tasks in \textsc{VGCure}, it still shows obvious improvement in both QA and FC samples for most tasks in \textsc{VGCure}. 
This demonstrates that the proposed method can improve the fundamental graph structure understanding capabilities of LVLMs, which leads to better performance on downstream tasks.
We also present a case on \textit{NR} to further understand the effectiveness of the proposed method, please refer to Appendix \ref{app:case_study} for detailed information.

\begin{figure}[t]
    \centering
    \includegraphics[width=\linewidth]{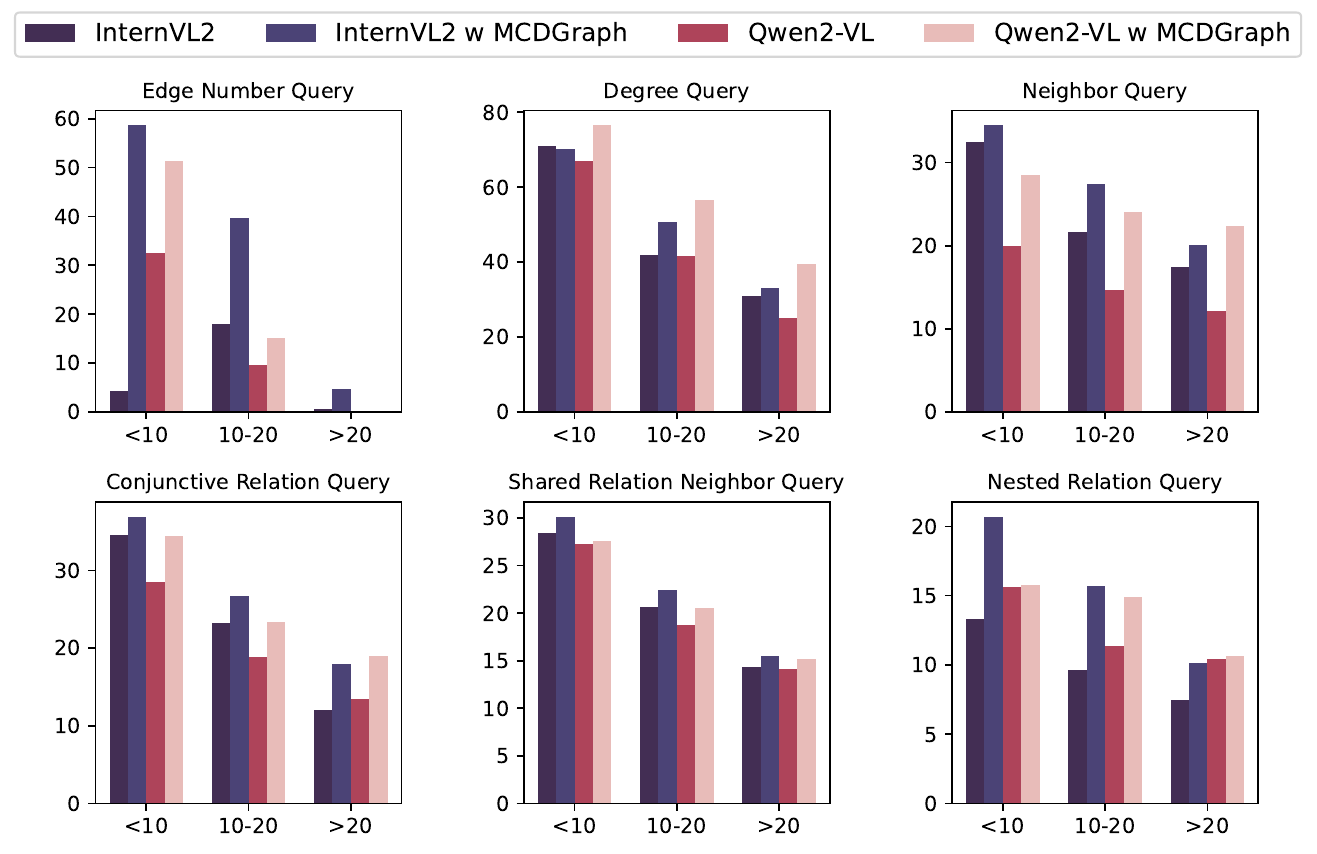}
     \vspace{-0.8cm}
    \caption{Model performance (F1/Acc) comparison on QA samples across representative tasks and edge ranges.}
    \label{fig:size_results_MCDGraph_qa_sampling}
\end{figure}

\noindent \textbf{Ablation Study} We also conduct ablation study on QA samples with Qwen2-VL and the results are shown in Tab.\ref{tab:ablation_study}. It can be observed that after removing different training tasks, although the LVLM may have a slight performance improvement on some tasks, it may lose certain capabilities causing its performance to plummet on some task (as evidenced by the red results in the Tab.\ref{tab:ablation_study}).
Meanwhile, all three self-supervised tasks enhance the LVLMs' ability to capture structural and relational information from different dimensions, complementing each other in order to comprehensively improve the LVLM's overall performance on all tasks.

\begin{table}[t]
   \centering
    \resizebox{\linewidth}{!}{
\begin{tabular}{lcccccccccc}
\toprule
\multirow{2.5}{*}{\textbf{Models}} & \multicolumn{4}{c}{\textbf{Understanding}}                           & \multicolumn{6}{c}{\textbf{Reasoning}}                                                           \\  \cmidrule(lr){2-5}  \cmidrule(lr){6-11}
                                 & \textbf{NNu} & \textbf{EN} & \textbf{DQ} & \textbf{NQ} & \textbf{NN} & \textbf{CR}  & \textbf{RA} & \textbf{SRN} & \textbf{NR} & \textbf{CP} \\
                                \midrule
\textbf{Qwen2-VL}        & 97.80         & 16.38       & 48.09       & 16.18       & 16.52       & 21.02       & 14.19       & 19.48        & 12.73         & 12.90         \\
\textbf{w MCDGraph }     & 98.34         & 25.92       & 60.94       & 25.44       & 13.32       & 26.14       & 13.14       & 20.74        & 14.44         & 11.95         \\
\textbf{- Masked }       & 99.02         & 32.11       & 57.75       & 22.84       & 14.15       & 23.29       & 15.32       & 19.15        & \textcolor{red}{5.01} & 24.61         \\
\textbf{- Contrastive}   & 99.64         & 28.70       & 64.02       & 25.35       & 15.16       & 25.13       & 11.83       & 21.11        & 15.20         & \textcolor{red}{5.33} \\
\textbf{- Description }  & \textcolor{red}{2.00} & 13.27       & 57.80       & 23.09       & 14.14       & 23.47       & 10.16       & 19.87        & 16.99         & \textcolor{red}{8.41} \\
\bottomrule
\end{tabular}}
\vspace{-0.2cm}
    \caption{Ablation study on across various tasks.}
    \label{tab:ablation_study}
\end{table}

\noindent \textbf{Performance on Varying Complexity} 
Fig.\ref{fig:size_results_MCDGraph_qa_sampling} illustrates the performance improvements of \textsc{MCDGraph} on LVLMs across varying graph complexities on six representative tasks in \textsc{VGCure}.
Fine-tuning with \textsc{MCDGraph} consistently improves performance across most tasks and complexity levels, demonstrating its its effectiveness in enhancing the LVLMs' ability to understand and reason over visual graphs. 
For simpler graphs, the improvement is smaller, where LVLMs already perform well, but as complexity increases, the performance gap between fine-tuned and baseline models becomes more pronounced, highlighting \textsc{MCDGraph}'s importance in handling more complex visual graphs.
Due to the space limit, the results for other dimensions are shown in Figs.\ref{fig:size_results_MCDGraph_qa}-\ref{fig:degree_results_MCDGraph_fc} in the Appendix, with conclusions similar to those above.

\smallskip

\noindent \textbf{Generalization of \textsc{MCDGraph}} 
To validate the generalization of our method, we regenerate 50 visual graphs with different \textbf{visual styles} and \textbf{naming conventions} of nodes and edges from those in \textsc{VGCure} for each graph structure. The details and results are presented in Appendix \ref{app:generalization}.

\begin{table}[t]
   \centering
    \resizebox{\linewidth}{!}{
\begin{tabular}{lccccccc}
\toprule
\multirow{2.5}{*}{\textbf{Model}} & \multicolumn{3}{c}{\textbf{VisionGraph }} & \multicolumn{4}{c}{\textbf{\textsc{FactKG} }}            \\
\cmidrule(lr){2-4}\cmidrule(lr){5-8}
                       & \textbf{Connect}   & \textbf{Cycle}  & \textbf{MaxFlow } & \textbf{Accuracy} & \textbf{Precision} & \textbf{Recall} & \textbf{F1}    \\
\midrule
\textbf{Qwen2-VL}               & 55.8      & 52.88  & 1.72       & 79.60    & 81.18     & 79.13  & 79.13 \\
\rowcolor{blue!10}\textbf{w \textsc{MCDGraph}}                & 53.37     & 52.88  & \textbf{5.17}       & \textbf{80.15}    & \textbf{81.71}     & \textbf{79.69}  & \textbf{79.71} \\
\textbf{InternVL2 }             & 46.9      & 52.88  & 6.9        & 79.41    & 80.93     & 78.95  & 78.95 \\
\rowcolor{blue!10}\textbf{w \textsc{MCDGraph}}               & \textbf{54.72}     & 52.88  & \textbf{8.62}       &\textbf{79.78}    & \textbf{81.34}     & \textbf{79.32}  & \textbf{79.33} \\
\bottomrule
\end{tabular}}
\vspace{-0.2cm}
        \caption{Model performance on downstream tasks.}
            \label{tab:downstream_results}
\end{table}

\subsection{Results on Downstream Reasoning Tasks}
To further demonstrate the scalability and applicability of our method, we evaluate \textsc{MCDGraph} on graph-related downstream reasoning tasks.

\smallskip

\noindent \textbf{VisionGraph} We first evaluate the performance of \textsc{MCDGraph} on three representative {graph theory problems} in VisionGraph \cite{li2024visiongraph}. As the results shown in Tab.\ref{tab:downstream_results}, \textsc{MCDGraph} generally improve the performance of LVLMs on these tasks, especially for the relatively difficult \textit{Maximum Flow task}. This confirms the effectiveness and scalability of our method.

\smallskip

\noindent \textbf{\textsc{FactKG}} We then evaluate \textsc{MCDGraph} on \textsc{FactKG} \cite{kim-etal-2023-factkg}, a knowledge graph-based fact verification dataset collected from \textbf{real-world data}. 
As Tab.\ref{tab:downstream_results} shows, \textsc{MCDGraph} gives consistently better performance than LVLMs on \textsc{FactKG}, suggesting that the proposed method can improve LVLMs in real-world graph-related tasks. 
Note that \textsc{FactKG} not only requires fundamental visual graph understanding and reasoning abilities but also relies on LVLMs' understanding of semantic and logical relationships between entities, which our method does not address. Therefore, the improvement of LVLMs is not as significant as that of the graph theory problems. We also evaluate our method on two \textbf{general vision reasoning tasks} and the results are available in Appendix \ref{app:vqa_results}.


\section{Related Work}
\noindent \textbf{Multimodal Benchmark for Graphs}
\citet{li2024visiongraph} and \citet{wei2024gita} introduce VisionGraph and GVLQA, respectively, for testing the problem-solving capabilities of LVLMs in graph theory. Both of them contain numerous synthetic visual graphs and complex graph theory problems.
Besides, \citet{ai-etal-2024-advancement} propose a novel instruction-following benchmark for multimodal graph understanding and reasoning, which contains a number of real-world graph images with diverse structures across various domains. 
However, these benchmarks focus on specific downstream tasks where LVLMs perform poorly. The goal of \textsc{VGCure} is to assess LVLMs' fundamental understanding and reasoning abilities on visual graphs to identify the reasons for their failures.

\smallskip

\noindent \textbf{Boosting LVLMs for Visual Graph Reasoning}
\citet{li2024visiongraph} propose a Description-Program-Reasoning (DPR) chain to enhance logical accuracy through graphical structure description and multi-step reasoning. \citet{wei2024gita} introduce GITA, an end-to-end framework integrating visual information into instruction-based graph reasoning. Additionally, \citet{deng2024graphvis} present GraphVis, which uses curriculum fine-tuning for training LVLMs on feature recognition and visual graph QA tasks.
Unlike current methods, our \textsc{MCDGraph} is a general-purpose, self-supervised approach that improves the fundamental understanding and reasoning of LVLMs on visual graphs, making it adaptable to most graph-related downstream tasks.

\smallskip

\noindent \textbf{Graph Benchmarks for GNNs} \citet{rozemberczki2021multi} construct Wikipedia-based graphs with pages as nodes and hyperlinks as edges. \citet{hu2020open} present realistic datasets spanning social, biological, molecular, code, and knowledge graphs. \citet{mernyei2020wiki} propose Wiki-CS, a GNN benchmark based on Computer Science articles. \citet{morris2020tudataset} release TUDataset, covering 120 datasets across multiple domains for graph classification and regression. To address homophily limitations, \citet{lim2021large} introduce large-scale, non-homophilous graph datasets. \citet{dwivedi2022long} propose the Long Range Graph Benchmark to evaluate models on long-range interaction reasoning tasks. Different from these graph benchmarking approach, the proposed VGCure focuses on exploring the fundamental graph understanding and reasoning capabilities of LVLMs, showcasing their potential to unify multimodal information processing through a unified visual learning paradigm.

\section{Conclusion}
This paper introduces \textsc{VGCure}, a comprehensive benchmark comprising 22 tasks to evaluate LVLMs' fundamental understanding and reasoning capabilities on visual graphs. Experiments on 14 LVLMs reveal significant limitations, especially in capturing structural information. To this end, we propose \textsc{MCDGraph}, a structure-aware self-supervised method to enhance open-source LVLMs' structure learning abilities. Extensive experiments validate the effectiveness of our method across a wide range of graph-related tasks.

\section*{Limitations}
\begin{itemize}
    \item \textbf{Complexity of Visual Graphs.} Due to the limitations of current LVLMs' performance on visual graph tasks, we restrict the number of nodes in the synthetic graph structure to between 7 and 15, potentially limiting the exploration and improvement of the LVLMs' performance on more complex visual graphs.
    \item \textbf{Experiments on Larger LVLMs.} Due to limited resources, the majority of our experiments are performed only on LVLMs with around 7B parameters, lacking performance evaluation and improvement of larger models with more parameters.
\end{itemize}

\section*{Acknowledgments}

We would like to thank the anonymous reviewers and meta-reviewer for their insightful suggestions. This work was partly supported by the National Natural Science Foundation of China (62406091, 62276077, 62125201, U23B2055, U24A20328, U24B20174, 62350710797), the National Science and Technology Innovation 2030 Major program  (2024ZD01NL00101), and Shenzhen Science and Technology Program (KQTD2024072910215406, ZDSYS20230626091203008).

\bibliography{main}

\appendix

\section{\textsc{VGCure} Construction}
\label{app:benchmark}
\subsection{Graphs Generation}
For synthetic graph structures, we use the NetworkX library for random generation and employ hyperparameters to control the expected macroscopic properties of each graph:
\begin{itemize}[itemsep=3pt,topsep=1pt,parsep=1pt,leftmargin=11pt]
    \item \textbf{ER}: This structure takes an edge probablity parameter $p$, which we choose randomly from $\{0.2, 0.3, 0.4\}$ during generation. 
    \item \textbf{BA}: This structure takes the parameter $m$, which denotes the number of edges to attach from a new node to existing nodes. We choose randomly from $\{2, 3\}$ during generation.
    \item \textbf{SFN}: For this structure, we use the default parameters provided by NetworkX except for the number of nodes during generation. 
    \item \textbf{SBM}: This structure takes the sizes of blocks $s$ and the density of edges going from the nodes of one group to nodes of another group $p$ as parameters. During generation, we set $s$ to $[n, m]$ and $p$ to $[[p_1, p_2], [p_2, p_3]]$, where $n$ and $m$ are a random integer from $[3, 7]$ and $[4, 8]$, respectively, and $p_1, p_2, p_3$ are all randomly selected from $\{0.2, 0.3, 0.4\}$.
    \item \textbf{Star}: This structure requires no parameters other than the number of nodes.
\end{itemize}

For all the above structures except SBM, the number of nodes during generation is an arbitrary integer in the range $[7, 15]$. 

\begin{figure*}[htbp]
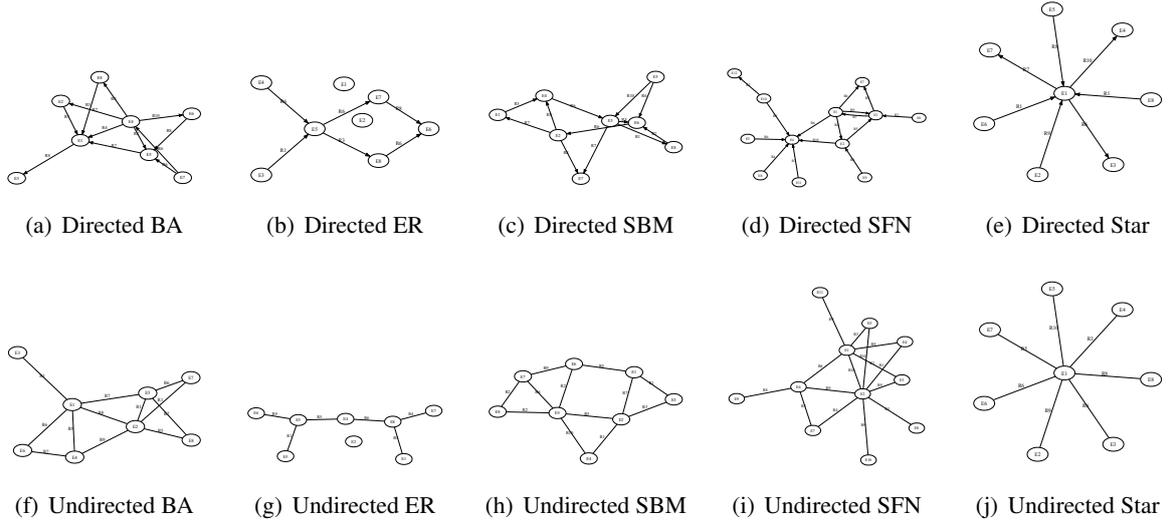

    \centering
    \subfigure[Directed BA]{
        \includegraphics[width=0.18\textwidth]{figures/6_ba_directed.pdf}
    }
    \subfigure[Directed ER]{
        \includegraphics[width=0.18\textwidth]{figures/7_er_directed.pdf}
    }
    \subfigure[Directed SBM]{
        \includegraphics[width=0.18\textwidth]{figures/25_sbm_directed.pdf}
    }
    \subfigure[Directed SFN]{
        \includegraphics[width=0.18\textwidth]{figures/1786_sfn_directed.pdf}
    }
    \subfigure[Directed Star]{
        \includegraphics[width=0.18\textwidth]{figures/13_star_directed.pdf}
    }

    \subfigure[Undirected BA]{
        \includegraphics[width=0.18\textwidth]{figures/16_ba_undirected.pdf}
    }
    \subfigure[Undirected ER]{
        \includegraphics[width=0.18\textwidth]{figures/6_er_undirected.pdf}
    }
    \subfigure[Undirected SBM]{
        \includegraphics[width=0.18\textwidth]{figures/6_sbm_undirected.pdf}
    }
    \subfigure[Undirected SFN]{
        \includegraphics[width=0.18\textwidth]{figures/1258_sfn_undirected.pdf}
    }
    \subfigure[Undirected Star]{
        \includegraphics[width=0.18\textwidth]{figures/17_star_undirected.pdf}
    }
    \caption{Examples of synthetic visual graphs.}
    \label{fig:example_graphs}
\end{figure*}

To anonymize the visual graph, we use the unique name E$x$ containing no information to name the nodes in the graph structure, where $x\in\{1,2,\dots,n\}$ and $n$ is the number of nodes in the graph. For edges, we choose a random identify from \{R1, R2, ... , R10\} to name them. The name can be repeated for each edge.
The examples of synthetic visual graph are shown in Fig.\ref{fig:example_graphs}.

\subsection{Tasks Generation}
To ensure the correctness of the generated samples, we first search for relevant paths in the given graph that satisfy the conditions of the task. Then the final QA samples and FC samples are generated based on the paths and corresponding templates. If no path exists, the generation of samples for the task is skipped.
The final statistics of \textsc{VGCure} and the example samples with undirected graph for each task are shown in Tab.\ref{tab:statistics} and Tab.\ref{tab:examples_undirected}, respectively.

\section{Experimental Setup for Evaluation}
\label{app:evaluation_experimental_setup}

\subsection{Prompts}
To facilitate the LVLMs to understand the content in the visual graph, we take a \textbf{visual graph description} in addition to the input during test.
\begin{itemize}[itemsep=3pt,topsep=1pt,parsep=1pt,leftmargin=11pt]
    \item \textbf{Directed Visual Graph Description}: The given image shows a graph where circles represent nodes, with the content inside indicating the node names. The arrowed lines connecting two nodes represent edges, and the content in the middle of the edges indicates the edge names.
    \item \textbf{Undirected Visual Graph Description}: The given image shows a graph where circles represent nodes, with the content inside being the node names. The lines connecting two nodes represent edges, and the content in the middle of the edges represents the edge names.
\end{itemize}
The complete prompt for QA samples are as follow:
\begin{itemize}[itemsep=3pt,topsep=1pt,parsep=1pt,leftmargin=11pt]
    \item \textbf{NN, CR, RA, SRN, NQ}: [Visual Graph Description] Answer the given questions based on the graph in the image.$\setminus$nQuestion: [question]$\setminus$nPlease provide the answer directly without the reasoning process and present your answer in the LIST format: [Entity1, Entity2, ...].
    \item \textbf{NR}: [Visual Graph Description] Answer the given questions based on the graph in the image.$\setminus$nQuestion: [question]$\setminus$nPlease provide the answer directly without the reasoning process and present your answer in the LIST format: [Relation1, Relation2, ...].
    \item \textbf{CP}: [Visual Graph Description] Answer the given questions based on the graph in the image.$\setminus$nQuestion: [question]$\setminus$nIf yes, please output all the shortest paths in the LIST Format and conclude your answer with "Yes. The shortest paths are [[Entity1, Entity2,...], [Entity3, Entity4,...], ...]". If no path exists, please answer "No".
    \item \textbf{NNu, EN, DQ}: [Visual Graph Description] Answer the given questions based on the graph in the image.$\setminus$nQuestion: [question]$\setminus$nPlease provide the answer directly without the reasoning process.
\end{itemize}
The complete prompt for FC samples are as follow:
\begin{itemize}[itemsep=3pt,topsep=1pt,parsep=1pt,leftmargin=11pt]
    \item {}[Visual Graph Description] Verify the truth of the given claim against the graph in the image.$\setminus$nClaim: [claim]$\setminus$nThe answer should contain only "True" or "False", without reasoning process.
\end{itemize}

\subsection{Evaluation Metrics}
For the QA samples of NQ, NN, CR, RA, SRN, and NR tasks, we use (macro-averaged) F1 score and Hits@1 as in the previous QA benchmarks~\cite{rajpurkar-etal-2016-squad, Zhang_Dai_Kozareva_Smola_Song_2018}. For the QA samples of CP task, we employ EM\_F1, which is the macro F1 score calculated based on the exact match between the predicted path and the ground truth path, and Label\_Acc, which measures the accuracy of the model’s prediction on whether a path exists or not. For the QA samples of NNu, EN and DQ tasks, we compute the accuracy between predicted answers and ground truth. For the FC samples of all tasks,
following~\citet{si-etal-2024-checkwhy, si-etal-2024-denoising}, 
we use macro F1 and accuracy as the metrics.

\begin{table*}[t]
   \centering
    \resizebox{\linewidth}{!}{
    \renewcommand\arraystretch{1.2}
\begin{tabular}{lcccccccccccccccc}
\toprule
\multicolumn{2}{c}{\multirow{2.2}{*}{\textbf{Structue Type}}}   &\multirow{2.2}{*}{\textbf{\# Graphs}}   & \multicolumn{12}{c}{\textbf{Number of QA Samples}}   & \multirow{2.2}{*}{\textbf{Avg. Nodes}} & \multirow{2.2}{*}{\textbf{Avg. Edges}}  \\
\cmidrule{4-15}
\multicolumn{2}{c}{}                                           & & \textbf{NN} & \textbf{CR} & \textbf{CN} & \textbf{RA} & \textbf{SRN} & \textbf{NR} & \textbf{CP} & \textbf{NNu} & \textbf{EN} & \textbf{DC} & \textbf{DQ} & \textbf{NQ} \\
\midrule
\multirow{2}{*}{BA}                & Directed      & 400    & 400  & 400  & 400  & 370  & 324  & 400  & 400  & 400  & 400  & 400  & 400  & 400  & 10.98                       & 20.99                       \\
                                   & Undirected    & 400    & 400  & 400  & 400  & 400  & 400  & 400  & 400  & 400  & 400  & 400  & 400  & 400  & 11.02                       & 21.04                       \\\midrule
\multirow{2}{*}{ER}                & Directed      & 400    & 387  & 377  & 377  & 304  & 229  & 387  & 400  & 400  & 400  & 400  & 400  & 400  & 11.06                       & 17.69                       \\
                                   & Undirected    & 400    & 396  & 395  & 396  & 346  & 346  & 395  & 400  & 400  & 400  & 400  & 400  & 400  & 11.00                       & 17.41                       \\\midrule
\multirow{2}{*}{SBM}               & Directed      & 400    & 399  & 392  & 392  & 314  & 399  & 400  & 253  & 400  & 400  & 400  & 400  & 400  & 10.99                       & 17.13                       \\
                                   & Undirected    & 400    & 399  & 399  & 399  & 399  & 400  & 371  & 371  & 400  & 400  & 400  & 400  & 400  & 11.04                       & 17.32                       \\\midrule
\multirow{2}{*}{SFN}               & Directed      & 400    & 400  & 400  & 400  & 331  & 149  & 400  & 400  & 400  & 400  & 400  & 400  & 400  & 10.95                       & 13.86                       \\
                                   & Undirected    & 400    & 400  & 400  & 400  & 400  & 400  & 379  & 379  & 400  & 400  & 400  & 400  & 400  & 11.05                       & 12.80                       \\\midrule
\multirow{2}{*}{Star}              & Directed      & 400    & 396  & 388  & 388  & 350  & 288  & 396  & 400  & 400  & 400  & 400  & 400  & 400  & 12.05                       & 11.05                       \\
                                   & Undirected    & 400    & 400  & 400  & 400  & 393  & 393  & 400  & 400  & 400  & 400  & 400  & 400  & 400  & 12.11                       & 11.11                       \\\midrule
\multirow{2}{*}{DBLP}              & Directed      & 200    & 196  & 192  & 192  & 140  & 196  & 200  & 145  & 200  & 200  & 200  & 200  & 200  & 7.85                        & 17.76                       \\
                                   & Undirected    & 200    & 200  & 200  & 200  & 173  & 173  & 200  & 200  & 200  & 200  & 200  & 200  & 200  & 7.85                        & 17.76                       \\\midrule
\multirow{2}{*}{Dbpedia}           & Directed      & 200    & 186  & 185  & 185  & 93   & 186  & 200  & 101  & 200  & 200  & 200  & 200  & 200  & 8.13                        & 11.36                       \\
                                   & Undirected    & 200    & 200  & 200  & 200  & 155  & 155  & 200  & 200  & 200  & 200  & 200  & 200  & 200  & 8.13                        & 11.36                       \\\midrule
\multirow{2}{*}{Openflights}       & Directed      & 100    & 100  & 100  & 100  & 65   & 100  & 100  & 63   & 100  & 100  & 100  & 100  & 100  & 5.51                        & 13.59                       \\
                                   & Undirected    & 100    & 100  & 100  & 100  & 90   & 90   & 100  & 100  & 100  & 100  & 100  & 100  & 100  & 5.51                        & 13.59                       \\\midrule
\multirow{2}{*}{PubChemQC}         & Directed      & 400    & 400  & 138  & 400  & 400  & 119  & 119  & 37   & 400  & 400  & 400  & 400  & 400  & 5.45                        & 4.76                        \\
                                   & Undirected    & 400    & 400  & 398  & 400  & 398  & 400  & 151  & 151  & 400  & 400  & 400  & 400  & 400  & 5.45                        & 4.76                        \\\midrule
\multirow{2}{*}{Social Network}    & Directed      & 300    & 262  & 254  & 254  & 134  & 262  & 300  & 124  & 300  & 300  & 300  & 300  & 300  & 7.57                        & 9.25                        \\
                                   & Undirected    & 300    & 299  & 297  & 299  & 238  & 238  & 297  & 300  & 300  & 300  & 300  & 300  & 300  & 7.57                        & 9.25                        \\\midrule
\multicolumn{2}{l}{Total}                          & 6400   & 6320 & 6015 & 6282 & 5493 & 5247 & 5795 & 5224 & 6400 & 6400 & 6400 & 6400 & 6400 & -                           & -                              \\
\bottomrule
\end{tabular}}
    \caption{Statistics of \textsc{VGCure} benchmark, where \textit{\# Graphs} represents the number of visual graphs, \textit{Avg.Nodes} and \textit{Avg.Edges} denote the average number of nodes and edges in the graph, respectively. For each task, the number of QA samples is the values in the table, except for CN and DC, which have no QA samples, and the number of FC samples is \textbf{twice} the value in the table.}
    \label{tab:statistics}
\end{table*}

\begin{table*}[ht]
   \centering
    \resizebox{\linewidth}{!}{
\begin{tabular}{p{1.5cm}ll}
\toprule
\multicolumn{1}{c}{\textbf{Task}} & \multicolumn{1}{c}{\textbf{QA sample}}                 & \multicolumn{1}{c}{\textbf{FC sample (Label)}}                                      \\
\midrule

\multicolumn{1}{c}{\multirow{2}{*}{\textbf{NNu}}}              & \multirow{2}{*}{\begin{tabular}[c]{@{}l@{}}Q: How many nodes are there in this graph?\\ A: 11\end{tabular}}                                                    & There are 11 nodes in this graph. (True)                                                                              \\ &      & There are 15 nodes in this graph. (False)   \\
\midrule
\multicolumn{1}{c}{\multirow{2}{*}{\textbf{EN}}}               & \multirow{2}{*}{\begin{tabular}[c]{@{}l@{}}Q: How many edges are there in this graph?\\ A: 15\end{tabular}}                                                    & There are 15 edges in this graph. (True)     \\   & & There are 19 edges in this graph. (True)     \\
\midrule
\multicolumn{1}{c}{\multirow{2}{*}{\textbf{DC}}}   & \multirow{2}{*}{-}    & This graph is an undirected graph. (True)    \\&    & This graph is a directed graph. (True)  \\
\midrule
\multicolumn{1}{c}{\multirow{2}{*}{\textbf{DQ}}}      & \multirow{2}{*}{\begin{tabular}[c]{@{}l@{}}Q: What is the degree of E7 in this graph?\\ A: 2\end{tabular}}                                                     & The degree of E7 in this graph is 2. (True)                                                                           \\  &    & The degree of E7 in this graph is 7. (False)    \\
\midrule
\multicolumn{1}{c}{\multirow{2}{*}{\textbf{NQ}}}              & \multirow{2}{*}{\begin{tabular}[c]{@{}l@{}}Q: Which nodes are neighbors of E6 in this graph?\\ A: {[}E1, E2, E7, E9{]}\end{tabular}}                                   & E7 is a neighbors of E6 in this graph. (True)                                                                     \\ &   & E10 is a neighbors of E6 in this graph. (False)      \\  
\midrule
\multicolumn{1}{c}{\multirow{2}{*}{\textbf{NN}}}               & \multirow{2}{*}{\begin{tabular}[c]{@{}l@{}}Q: Which entities are connected to the entity that has R10 with E2 via R9?\\ A: {[}E11, E4{]}\end{tabular}}                            & E4 is connected to the entity that has R10 with E2 via R9. (True)                                                                       \\  &        & E3 is connected to the entity that has R10 with E2 via R9. (False)    \\
\midrule
\multicolumn{1}{c}{\multirow{2}{*}{\textbf{CR}}}               & \multirow{2}{*}{\begin{tabular}[c]{@{}l@{}}Q: Which entities are connected to E8 via R3 as well as connected E1 via R10?\\ A: {[}E2{]}\end{tabular}}                                     & E2 is connected to E8 via R3 as well as connected E1 via R10. (True)                                                                            \\ &      & E3 is connected to E8 via R3 as well as connected E1 via R10. (False)       \\
\midrule
\multicolumn{1}{c}{\multirow{2}{*}{\textbf{CN}}}               & {\multirow{2}{*}{-}}                                                                                                                         & E5 and E2 share a common neighbor. (True)    \\  & \multicolumn{1}{c}{}      & E10 and E11 share a common neighbor. (False)  \\
\midrule
\multicolumn{1}{c}{\multirow{2}{*}{\textbf{RA}}}               & \multirow{2}{*}{\begin{tabular}[c]{@{}l@{}}Q: Which entities are connected to E1 via the same relation between E11 and E1?\\ A: {[}E4{]}\end{tabular}}           & E4 is connected to E1 via the same relation between E11 and E1. (True)          \\ &      & E2 is connected to E1 via the same relation between E11 and E1. (False)       \\
\midrule
\multicolumn{1}{c}{\multirow{2}{*}{\textbf{SRN}}}              & \multirow{2}{*}{\begin{tabular}[c]{@{}l@{}}Q: Which entities are both connected to E2 via R10?\\ A: {[}E1, E5{]}\end{tabular}}                                              & E5 and E1 both connected to E2 via R10. (True)                                                                                 \\   &      & E5 and E3 are both connected to E2 via R10. (False)         \\
\midrule
\multicolumn{1}{c}{\multirow{2}{*}{\textbf{NR}}}               & \multirow{2}{*}{\begin{tabular}[c]{@{}l@{}}Q: What is the relation between E2 and the entity that is connected to E6 via R8? \\ A: {[}R10{]}\end{tabular}}                        & The relation between E2 and the entity that is connected to E6 via R8 is R10. (True)                                                                       \\ &    & The relation between E2 and the entity that is connected to E6 via R8 is R4. (False)    \\
\midrule
\multicolumn{1}{c}{\multirow{2}{*}{\textbf{CP}}}               & \multirow{2}{*}{\begin{tabular}[c]{@{}l@{}}Q: Is there a path between E5 and E3?\\ A: Yes. The shortest paths are {[}{[}E5, E1, E3{]}, {[}E5, E2, E3{]}{]}\end{tabular}} & [E5, E1, E3] is one of the shortest path between E5 and E3. (True)                                                                               \\   &      & [E5, E11, E2, E3] is one of the shortest path between E5 and E3. (False)   \\
\bottomrule
\end{tabular}}
\caption{Examples with undirected graph for each task in \textsc{VGCure}. These samples all correspond to the SFN graph shown in Fig.\ref{fig:example_graphs}(i).}
\label{tab:examples_undirected}
\end{table*}

\begin{table*}[t]
   \centering
    \resizebox{\linewidth}{!}{
\begin{tabular}{llc}
\toprule
\textbf{Error   Type}              & \multicolumn{1}{c}{\textbf{QA samples}}                                                                                                                                                                            & \textbf{Visual Graph} \\
\midrule
\textbf{Relation Misunderstanding} & \begin{tabular}[c]{@{}l@{}}Q: What is the relation from the entity that is R5 of E3 to E1?\\ A: {[}R1{]}\\ P: {[}\textcolor[RGB]{192,0,0}{R6}{]}\end{tabular}                                                                                & Fig.\ref{fig:error_graphs}(a)             \\\midrule
\textbf{Complexity Limitation}     & \begin{tabular}[c]{@{}l@{}}Q: What is the relation from the entity that is R7 of E1 to E7?\\ A: {[}R5{]}\\ P: {[}\textcolor[RGB]{192,0,0}{R9}{]}\end{tabular}                                                                                & Fig.\ref{fig:error_graphs}(b)             \\\midrule
\textbf{Structural Hallucination}  & \begin{tabular}[c]{@{}l@{}}Q: Is there a path between E1 and E3?\\ A: Yes. The shortest paths are {[}{[}E1, E6, E3{]}{]}\\ P: Yes. The shortest paths are \textcolor[RGB]{192,0,0}{{[}E1, E3{]}} and \textcolor[RGB]{192,0,0}{{[}E1, E4, E3{]}}.\end{tabular}          & Fig.\ref{fig:error_graphs}(c)             \\\midrule
\textbf{Entity-based Answering}  & \begin{tabular}[c]{@{}l@{}}Q: Which entities are R5 of \textcolor[RGB]{192,0,0}{E2} as well as R1 of \textcolor[RGB]{192,0,0}{E1}?\\ A: {[}E11{]}\\ P: {[}\textcolor[RGB]{192,0,0}{E2}, \textcolor[RGB]{192,0,0}{E1}{]}\end{tabular}                                                                                       & Fig.\ref{fig:error_graphs}(b)             \\\midrule
\textbf{Off-target Generation}     & \begin{tabular}[c]{@{}l@{}}Q: What is the relation between E2 and the entity that is connected to E4 via R9?\\ A: {[}R2{]}\\ P: {[}\textcolor[RGB]{192,0,0}{Edge}, \textcolor[RGB]{192,0,0}{Node}{]}\end{tabular}                                                      & Fig.\ref{fig:error_graphs}(d)             \\\midrule
\textbf{Format Error}              & \begin{tabular}[c]{@{}l@{}}Q: Is there a path between E4 and E1?\\ A: Yes. The shortest paths are {[}{[}E4, E3, E2, E1{]}{]}\\ P: Yes. The shortest paths are {[}{[}E4, \textcolor[RGB]{192,0,0}{R9}, E3, \textcolor[RGB]{192,0,0}{R2}, E2, \textcolor[RGB]{192,0,0}{R6}, E1{]}{]}.\end{tabular} & Fig.\ref{fig:error_graphs}(d)     \\
\bottomrule
\end{tabular}}
\caption{Examples of each error type, where \textit{Q} denotes question, \textit{A} denotes gold answer and \textit{P} denotes prediction.}
\label{tab:error_type}
\end{table*}

\begin{figure}[t]
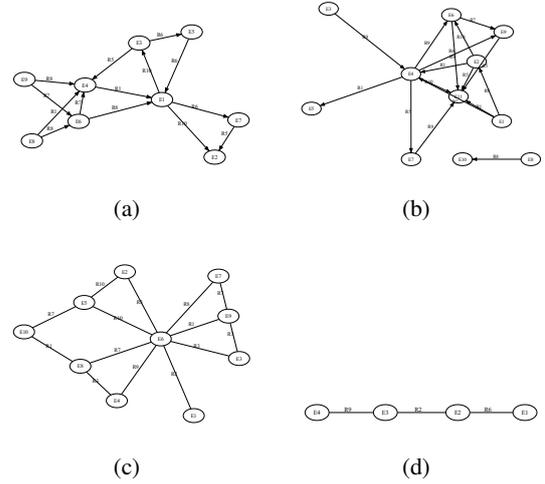

    \centering
    \subfigure[]{
        \includegraphics[width=0.22\textwidth]{figures/error_type_1.pdf}
    }
    \subfigure[]{
        \includegraphics[width=0.22\textwidth]{figures/error_type_2_4.pdf}
    }


    \subfigure[]{
        \includegraphics[width=0.22\textwidth]{figures/error_type_3.pdf}
    }
    \subfigure[]{
        \includegraphics[width=0.22\textwidth]{figures/error_type_5_6.pdf}
    }
    \caption{Visual graphs corresponding to the examples of error type.}
    \label{fig:error_graphs}
\end{figure}

\section{Why LVLMs Fail on Fundamental Tasks?}
\subsection{Comparison with Backbone LLMs}
\label{app:text_comparison}
To compare the performance of the LVLMs with the corresponding LLMs on the \textsc{VGCure} task, we first randomly select 1000 samples for each task and convert the corresponding visual graphs into text triples to construct the text version of the samples, and then evaluate them with the corresponding backbone LLMs.

The complete prompts for QA samples are as follows:
\begin{itemize}[itemsep=3pt,topsep=1pt,parsep=1pt,leftmargin=11pt]
    \item \textbf{NN, CR, RA, SRN, NQ}: Given a set of triples representing an directed/undirected graph, where each triple denotes [Node, Edge, Node], answer the given questions based on the graph.$\setminus$nTriples: [Triples of visual graph]$\setminus$nQuestion: [question]$\setminus$nPlease provide the answer directly without the reasoning process and present your answer in the LIST format: [Entity1, Entity2, ...].
    \item \textbf{NR}:  Given a set of triples representing an directed/undirected graph, where each triple denotes [Node, Edge, Node], answer the given questions based on the graph.$\setminus$nTriples: [Triples of visual graph]$\setminus$nQuestion: [question]$\setminus$nPlease provide the answer directly without the reasoning process and present your answer in the LIST format: [Relation1, Relation2, ...].
    \item \textbf{NNu, EN, DQ}:  Given a set of triples representing an directed/undirected graph, where each triple denotes [Node, Edge, Node], answer the given questions based on the graph.$\setminus$nTriples: [Triples of visual graph]$\setminus$nQuestion: [question]$\setminus$nPlease provide the answer directly without the reasoning process.
\end{itemize}
The complete prompt for FC samples are as follow:
\begin{itemize}[itemsep=3pt,topsep=1pt,parsep=1pt,leftmargin=11pt]
    \item {}[Visual Graph Description] Given a set of triples representing an directed graph, where each triple denotes [Node, Edge, Node], verify the truth of the given claim against the graph.$\setminus$nTriples: [Triples of visual graph]$\setminus$nClaim: [claim]$\setminus$nThe answer should contain only "True" or "False", without reasoning process.
\end{itemize}

The \textit{Connective Path Query (CP)} task is ignored here due to LLMs' poor instruction-following ability on this task.

\section{Experimental Setup for Training}
\label{app:training_experimental_setup}
\subsection{Training Samples}
\paragraph{Graphs Generation}
For Masked Graph Infilling and Graph Description task, we use the same synthetic visual graph generation strategy as \textsc{VGCure}. As for the Contrastive Graph Discrimination, in order to reduce the difficulty, we limited the number of nodes per graph structure to $[4, 8]$ during generation.

\paragraph{Task Instruction}
To increase the diversity of samples, we designed various instructions with similar semantics for each task in the \textsc{MCDGraph}.
\begin{itemize}[itemsep=3pt,topsep=1pt,parsep=1pt,leftmargin=11pt]
    \item \textbf{Masked Graph Infilling}
        \begin{itemize}[itemsep=3pt,topsep=1pt,parsep=1pt,leftmargin=9pt]
            \item Using the given set of triples, predict the word that should fill the [MASK] position in the knowledge graph in the image.
            \item Based on the provided triples, determine the correct word to complete the [MASK] position in the knowledge graph shown in the image.
            \item Given the set of triples, predict the word that should be placed in the [MASK] position within the knowledge graph in the image.
            \item Use the given triples to predict the appropriate word for the [MASK] position in the knowledge graph depicted in the image.
            \item Using the set of triples, identify the word that should fill the [MASK] position in the knowledge graph in the image.
            \item Based on the set of triples, predict the correct word for the [MASK] position in the knowledge graph in the image.
            \item Given the triples, predict the word that fits the [MASK] position in the knowledge graph present in the image.
            \item Using the triples provided, determine the word that should be used to fill the [MASK] position in the knowledge graph in the image.
            \item Predict the word that should occupy the [MASK] position in the knowledge graph in the image, based on the given triples.
            \item Using the provided triples, identify the word that should complete the [MASK] position in the knowledge graph in the image.
        \end{itemize}
    \item \textbf{Contrastive Graph Discrimination}
        \begin{itemize}[itemsep=3pt,topsep=1pt,parsep=1pt,leftmargin=9pt]
            \item Determine whether Graph A and Graph B in the given image are identical.
            \item Assess if Graph A and Graph B depicted in the image are equivalent.
            \item Evaluate whether the structures of Graph A and Graph B in the provided image match.
            \item Identify if there are any differences between Graph A and Graph B in the shown image.
            \item Check if Graph A and Graph B illustrated in the image are the same.
            \item Analyze the image to determine if Graph A is identical to Graph B.
            \item Investigate whether Graph A and Graph B in the given image are congruent.
            \item Examine the provided image to see if Graph A and Graph B are equivalent.
            \item Compare Graph A and Graph B in the image to establish their similarity.
            \item Confirm if Graph A and Graph B presented in the image are indistinguishable.
        \end{itemize}
    \item \textbf{Graph Description}
        \begin{itemize}[itemsep=3pt,topsep=1pt,parsep=1pt,leftmargin=9pt]
            \item Describe the given graph, including the number of nodes, the number of edges, and the names of all the nodes.
            \item Provide a description of the given graph, specifying the number of nodes, edges, and listing all the node names.
            \item Analyze the given graph by stating the number of nodes, edges, and enumerating the names of all the nodes.
            \item Summarize the graph by detailing the number of nodes, edges, and listing the names of each node.
            \item Explain the graph, including the count of nodes and edges, and provide the names of all the nodes.
            \item Describe the graph, indicating how many nodes and edges it contains, and listing all the node names.
            \item Provide an overview of the graph, mentioning the number of nodes, edges, and the names of all nodes.
            \item Characterize the given graph, noting the number of nodes, edges, and listing all the node names.
            \item Detail the structure of the given graph, including node and edge counts, and providing a list of all node names.
            \item Give a description of the graph, including the total number of nodes, edges, and the names of all the nodes.
        \end{itemize}
\end{itemize}
Similar to \textsc{VGCure}, we include a visual graph description in input as well. Thus, the complete \textbf{task instruction} $I$ for each training sample in \textsc{MCDGraph} is ``[Visual Graph Description] [Instruction]''.

\paragraph{Number of Samples} For Masked Graph Infilling task, we generate \textbf{10,000 samples}, with half of the samples masking nodes and the other half masking edges. For Contrastive Graph Discrimination tasks, \textbf{5,000 samples}, where each sample consists of two visual graphs, are generated automatically. Similarly, the Graph Description task also contains \textbf{5,000 samples}, each corresponding to a unique visual graphs.

\begin{table}[t]
   \centering
    \resizebox{\linewidth}{!}{
\begin{tabular}{lccccc}
\toprule
\textbf{Model}     & \multicolumn{1}{l}{\textbf{Lora\_rank}} & \multicolumn{1}{l}{\textbf{Lora\_ alpha}} & \multicolumn{1}{l}{\textbf{Global Batch Size}} & \multicolumn{1}{l}{\textbf{Learning rate}} & \multicolumn{1}{l}{\textbf{Epoch}} \\
\midrule
\textbf{Qwen2-VL}  & 8                                       & 16                                        & 64                                             & 1e-4                                       & 5                                  \\
\textbf{InternVL2} & 64                                      & 128                                       & 64                                             & 4e-5                                       & 1                              \\
\bottomrule
\end{tabular}}
\caption{Hyperparameters for training}
\label{tab:hyperparameters}
\end{table}

\subsection{Implementation Details}
For Qwen2-VL, we employ the LoRA-based supervised fine-tuning scripts provided by LLaMA-Factory\footnote{\url{https://github.com/hiyouga/LLaMA-Factory}}. For InternVL2, we perform LoRA-based fine-tuning based on the code and documentation provided officially\footnote{\url{https://github.com/OpenGVLab/InternVL}}. The hyperparameters used for training are shown in Tab.\ref{tab:hyperparameters}. All the experiments are finished on 4 A100 GPUs with 80GB memory.

\begin{table}[t]
 \centering
    \resizebox{0.85\linewidth}{!}{
\begin{tabular}{lcc}
\toprule
\textbf{Models}     & \textbf{ScienceQA} & \textbf{AOKVQA} \\
\midrule
\textbf{Qwen2-VL}   & 85.13              & 84.37           \\
\rowcolor{blue!10}\textbf{w \textsc{MCDGraph}} & 81.31              & 83.67           \\
\textbf{InternVL2}  & 97.07              & 85.41           \\
\rowcolor{blue!10}\textbf{w \textsc{MCDGraph}} & 96.88              & 84.45          \\ \bottomrule
\end{tabular}}
\caption{Model performance (Acc) on general VQA tasks.}
\label{tab:vqa_results}
\end{table}

\section{Generalization of \textsc{MCDGraph}}
\label{app:generalization}
\subsection{Impact on Visual Styles}

\begin{figure}[t]
    \centering
    \includegraphics[width=\linewidth]{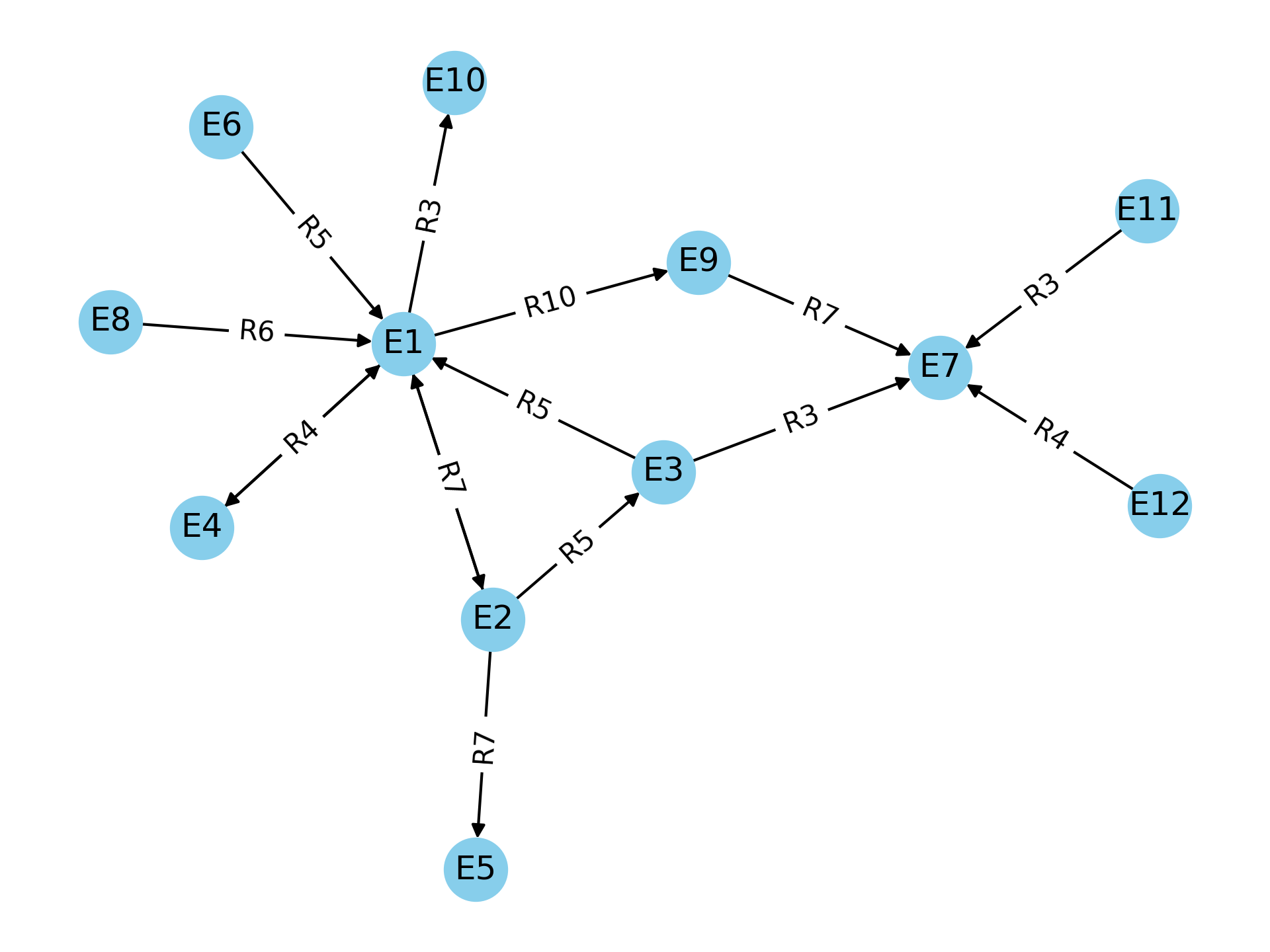}
    \caption{An example of visual graph with different style for the experimental results in Tab.\ref{tab:visual_style_results}}
    \label{fig:graph_networkx1}
\end{figure}

\begin{figure}[t]
    \centering
    \includegraphics[width=\linewidth]{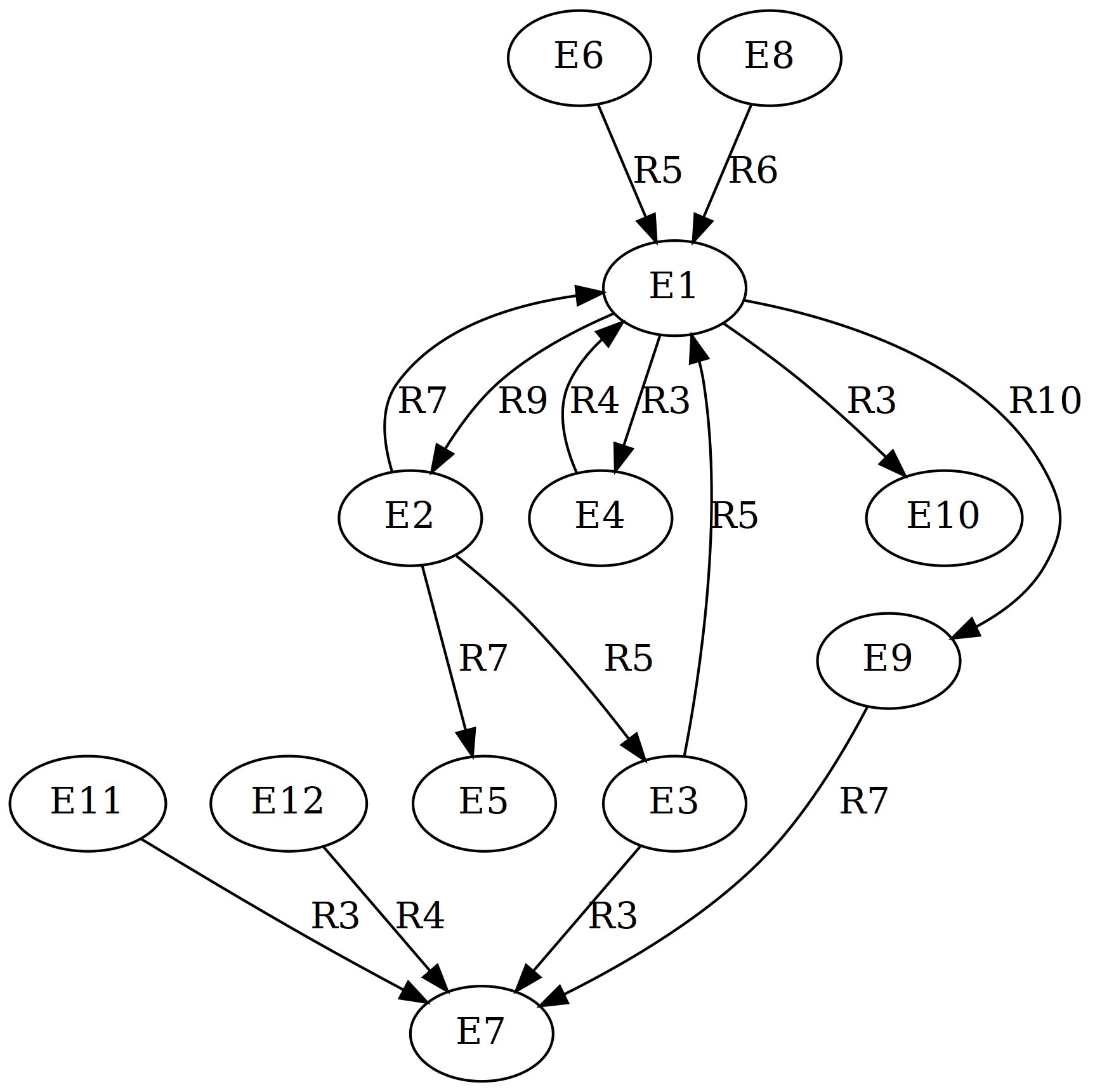}
    \caption{An example of visual graph with different style for the experimental results in Tab.\ref{tab:visual_style_1_results}}
    \label{fig:graph_networkx2}
\end{figure}

To validate the generalization of our method, we employ \textit{NetworkX} and \textit{Matplotlib} to regenerate 50 visual graphs with different visual styles from those in \textsc{VGCure} for each graph structure to explore the impact of visual graph styles. The examples of \textbf{the same} visual graph with different styles are illustrated in Fig.\ref{fig:graph_networkx1} and Fig.\ref{fig:graph_networkx2}.

As the results shown in Tabs.\ref{tab:visual_style_results} and \ref{tab:visual_style_1_results}, although LVLMs never encounter the different style of visual graph during fine-tuning, our method can still improve the performance of the LVLMs on almost all tasks. 
This demonstrates the ability of our method to enhance LVLMs’ ability in capturing the structural information in visual graphs with excellent generalization. In addition, compared to Tab.\ref{tab:MCDGraph_results}, it can be noticed that the experimental results before and after the change of style are similar, which indicates that the style itself has no effect on the evaluation of LVLMs.

\subsection{Impact of Naming Conventions}

To explore the impact of altering the naming conventions, we regenerated the visual graphs and samples using the following new naming conventions.
\begin{itemize}[itemsep=3pt,topsep=1pt,parsep=1pt,leftmargin=11pt]
    \item \textbf{Node-Edge}: The nodes in the visual graph are renamed \textbf{Node$x$}, where $x\in\{1,2,\dots,n\}$ and $n$ is the number of nodes in the graph, and the edges are renamed \textbf{Edge$y$}, where $y\in\{1,2,\dots,10\}$.
    \item \textbf{Name-R}: The nodes in the visual graph are renamed to a simple and common \textbf{human name} without any semantic bias, like ``John'', ``Jane'', ``Mike'', ``Mary'', etc. The names of the edges remain as they are, i.e., R1, R2, ..., R10.
\end{itemize}

The results are shown in Tabs.\ref{tab:naming_conventions_node_edge} and \ref{tab:naming_conventions_name_r}. We can observe that after altering the naming conventions, LVLMs continue to show similar trends on most of tasks. Therefore, the experimental analysis in the main text still holds.
Meanwhile, the performance of LVLMs decreases on most of the tasks when confronted with different names. This might be due to the fact that the new naming conventions gives longer names to nodes and edges and recognizing these information increases the difficulty of the task.
It is worth noting that in the face of the new naming conventions, our \textsc{MCDGraph} still improves the performance of LVLMs on most tasks, although the fine-tuning still uses the original naming conventions. This strongly demonstrates the effectiveness as well as the generalization of our method.

\section{Results on General VQA Tasks}
\label{app:vqa_results}

We evaluate our \textsc{MCDGraph} on two general VQA tasks, i.e., ScienceQA~\cite{NEURIPS2022_11332b6b} and AOKVQA~\cite{schwenk2022okvqa}. According to the results shown in Tab.\ref{tab:vqa_results}, we can observe that after MCDGraph, LVLMs perform worse on these two VQA tasks. This is due to the fact that our \textsc{MCDGraph} is proposed for visual graph understanding and reasoning tasks, which are very different from general VQA tasks. And it is worth noting that our method does not overly compromise the LVLMs' performance on general VQA tasks, which we consider acceptable.

\begin{figure}[t]
    \centering
    \includegraphics[width=\linewidth]{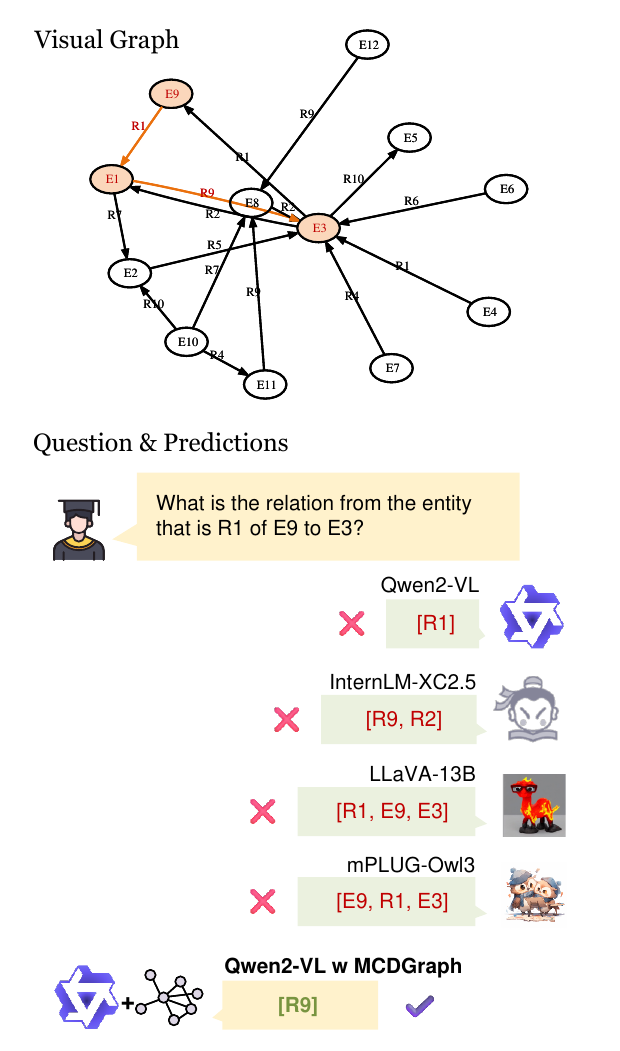}
    \caption{A case of Nested Relation Query task.}
    \label{fig:case_study}
\end{figure}

\section{Case Study}
\label{app:case_study}
Fig.\ref{fig:case_study} illustrates a case on \textit{Nested Relation Query} task. We can observe that Qwen2-VL, LLaVA-13B and mPLUG-Owl3 all make the error of \textbf{Entity-based Answering} when confronted with this question, i.e., using ``R1'', ``E9'', and ``E3'' mentioned in the question as the generated answer. InternLM-XC2.5 makes the error of \textbf{Relation Misunderstanding}, i.e., the edge pointing from E3 to E1 is also used as the answer. However, Qwen2-VL after applying \textsc{MCDGraph} can answer this question correctly. This demonstrates that the proposed method can improve the fundamental graph structure understanding of LVLMs, thus avoiding the occurrence of the previously mentioned errors.

\begin{table*}[t]

   \centering
    \resizebox{0.9\linewidth}{!}{
\begin{tabular}{l|cccccccccccc}
\toprule
\multirow{2.5}{*}{\textbf{Models}} & \multicolumn{5}{c}{\textbf{Understanding}}                           & \multicolumn{7}{c}{\textbf{Reasoning}}                                                           \\  \cmidrule(lr){2-6}  \cmidrule(lr){7-13}
                                 & \textbf{NNu} & \textbf{EN} & \textbf{DC} & \textbf{DQ} & \textbf{NQ} & \textbf{NN} & \textbf{CR} & \textbf{CN} & \textbf{RA} & \textbf{SRN} & \textbf{NR} & \textbf{CP} \\
                                 \midrule
                                 & \multicolumn{12}{c}{\textbf{QA Samples}}         \\
                                \midrule
\textbf{Qwen2-VL}                & 98.20 & 18.10 & - & 50.70 & 18.64 & 15.66 & 22.66 & - & 14.83 & 20.45 & 16.95 & 11.04      \\
\rowcolor{blue!10}\textbf{w \textsc{MCDGraph}}              & 99.40 $\uparrow$  & 40.20 $\uparrow$ & - & 56.80 $\uparrow$ & 28.44 $\uparrow$ & 14.64 & 27.02 & - & 13.69 & 22.22 $\uparrow$ & 14.73 & 11.99 $\uparrow$      \\
\textbf{InternVL2}              & 73.70 & 8.60  & - & 48.40 & 26.33 & 18.26 & 26.19 & - & 9.85  & 21.75 & 11.20 & 1.60     \\
\rowcolor{blue!10}\textbf{w \textsc{MCDGraph}}              & 95.80 $\uparrow$ & 35.70 $\uparrow$ & - & 52.90 $\uparrow$ & 28.89 $\uparrow$ & 19.90 $\uparrow$ & 29.24 $\uparrow$ & - & 11.73 $\uparrow$ & 23.23 $\uparrow$ & 18.04 $\uparrow$ & $\uparrow$ 14.49   \\
\midrule
                                 & \multicolumn{12}{c}{\textbf{FC Samples}}   \\\midrule
\textbf{Qwen2-VL}               & 71.37                   & 61.78                  & 86.53                  & 73.82                  & 73.03                  & 48.00                  & 77.35                  & 39.98                  & 36.04                  & 60.22                   & 45.31                  & 42.93     \\
\rowcolor{blue!10} \textbf{w \textsc{MCDGraph}}             & 88.45 $\uparrow$                   & 63.14   $\uparrow$                & 94.43      $\uparrow$             & 76.26    $\uparrow$               & 78.59  $\uparrow$                 & 63.17  $\uparrow$                 & 82.63  $\uparrow$                 & 53.78    $\uparrow$               & 54.33  $\uparrow$                 & 66.26  $\uparrow$                  & 65.95 $\uparrow$                  & 56.30  $\uparrow$                 \\
\textbf{InternVL2}             & 75.43                   & 34.80                  & 86.85                  & 63.68                  & 74.52                  & 61.12                  & 74.23                  & 47.22                  & 33.48                  & 55.85                   & 55.32                  & 53.15               \\
\rowcolor{blue!10}\textbf{w \textsc{MCDGraph}}              & 83.82  $\uparrow$                  & 70.14  $\uparrow$                 & 83.48                  & 60.77                  & 79.69  $\uparrow$                 & 71.39  $\uparrow$                 & 78.30  $\uparrow$                 & 52.38  $\uparrow$                 & 51.24  $\uparrow$                 & 59.83  $\uparrow$                  & 68.58  $\uparrow$                 & 44.73                     \\
\bottomrule
\end{tabular}}
\vspace{-0.2cm}
        \caption{Model performance (Acc/F1/EM\_F1 for QA and F1 for FC) on various tasks with different visual styles. The example of the corresponding visual style is shown in Fig.\ref{fig:graph_networkx1}}
            \label{tab:visual_style_results}
\end{table*}

\begin{table*}[t]

   \centering
    \resizebox{0.95\linewidth}{!}{
\begin{tabular}{l|cccccccccccc}
\toprule
\multirow{2.5}{*}{\textbf{Models}} & \multicolumn{5}{c}{\textbf{Understanding}}                           & \multicolumn{7}{c}{\textbf{Reasoning}}                                                           \\  \cmidrule(lr){2-6}  \cmidrule(lr){7-13}
                                 & \textbf{NNu} & \textbf{EN} & \textbf{DC} & \textbf{DQ} & \textbf{NQ} & \textbf{NN} & \textbf{CR} & \textbf{CN} & \textbf{RA} & \textbf{SRN} & \textbf{NR} & \textbf{CP} \\
                                 \midrule
                                 & \multicolumn{12}{c}{\textbf{QA Samples}}         \\
                                \midrule
\textbf{Qwen2-VL}                & 95.30 & 20.80 & -  & 40.00 & 13.95 & 8.73  & 19.04 & -  & 11.73 & 14.89 & 13.16 & 3.17    \\
\rowcolor{blue!10}\textbf{w \textsc{MCDGraph}}              & 98.80 $\uparrow$ & 17.30 & -  & 50.50 $\uparrow$ & 22.33  $\uparrow$ & 11.56  $\uparrow$ & 21.92  $\uparrow$ & -  & 11.05 & 17.84  $\uparrow$ & 11.60 & 0.12   \\
\textbf{InternVL2}             & 85.20 & 17.90 & -  & 39.60 & 24.21 & 19.99 & 22.26 & -  & 8.06  & 19.46 & 12.03 & 2.25     \\
\rowcolor{blue!10}\textbf{w \textsc{MCDGraph}}            & 97.20  $\uparrow$ & 35.70  $\uparrow$ & -  & 43.00  $\uparrow$ & 26.46  $\uparrow$ & 19.12 & 23.54  $\uparrow$ & -  & 11.00  $\uparrow$ & 20.45  $\uparrow$ & 14.80  $\uparrow$ & 12.11  $\uparrow$  \\
\midrule
                                 & \multicolumn{12}{c}{\textbf{FC Samples}}   \\\midrule
\textbf{Qwen2-VL}              & 70.88 & 68.06 & 66.94 & 57.90 & 65.09 & 43.62 & 67.32 & 34.28 & 34.79 & 53.23 & 42.54 & 42.89   \\
\rowcolor{blue!10} \textbf{w \textsc{MCDGraph}}             & 76.36  $\uparrow$ & 58.47  & 85.20  $\uparrow$ & 74.38  $\uparrow$ & 73.69 $\uparrow$  & 56.97  $\uparrow$ & 75.05 $\uparrow$  & 48.68  $\uparrow$ & 53.84 $\uparrow$  & 60.99  $\uparrow$ & 60.44 $\uparrow$  & 56.93 $\uparrow$  \\
\textbf{InternVL2}             & 68.42 & 38.35 & 91.84 & 64.58 & 65.45 & 62.91 & 72.42 & 48.17 & 34.60 & 57.88 & 57.77 & 52.48       \\
\rowcolor{blue!10}\textbf{w \textsc{MCDGraph}}             & 69.45 $\uparrow$  & 69.11  $\uparrow$ & 84.10 & 62.09 & 74.68  $\uparrow$ & 71.52  $\uparrow$ & 75.19  $\uparrow$ & 52.97  $\uparrow$ & 50.89  $\uparrow$ & 61.00  $\uparrow$ & 66.44  $\uparrow$ & 43.19               \\
\bottomrule
\end{tabular}}

        \caption{Model performance (Acc/F1/EM\_F1 for QA and F1 for FC) on various tasks with different visual styles. The example of the corresponding visual style is shown in Fig.\ref{fig:graph_networkx2}}
            \label{tab:visual_style_1_results}
\end{table*}

\begin{table*}[ht]

   \centering
    \resizebox{0.95\linewidth}{!}{
\begin{tabular}{l|cccccccccccc}
\toprule
\multirow{2.5}{*}{\textbf{Models}} & \multicolumn{5}{c}{\textbf{Understanding}}                           & \multicolumn{7}{c}{\textbf{Reasoning}}                                                           \\  \cmidrule(lr){2-6}  \cmidrule(lr){7-13}
                                 & \textbf{NNu} & \textbf{EN} & \textbf{DC} & \textbf{DQ} & \textbf{NQ} & \textbf{NN} & \textbf{CR} & \textbf{CN} & \textbf{RA} & \textbf{SRN} & \textbf{NR} & \textbf{CP} \\
                                 \midrule
                                 & \multicolumn{12}{c}{\textbf{QA Samples}}         \\
                                \midrule
\textbf{Qwen2-VL}               & 94.30        & 18.10       & -                    & 48.40       & 17.55       & 11.34       & 13.72       & -                    & 10.10       & 11.82        & 5.34        & 10.63     \\
\rowcolor{blue!10}\textbf{w \textsc{MCDGraph}}             & 94.10        & 16.00       & -                    & 63.50 $\uparrow$      & 17.25       & 8.50        & 14.56 $\uparrow$      & -                    & 8.29        & 12.81  $\uparrow$      & 12.71 $\uparrow$      & 13.00 $\uparrow$  \\
\textbf{InternVL2}             & 65.61        & 19.15       & -                    & 47.94       & 17.45       & 12.11       & 15.75       & -                    & 8.87        & 13.54        & 8.25        & 8.12      \\
\rowcolor{blue!10}\textbf{w \textsc{MCDGraph}}            & 89.30 $\uparrow$       & 34.60 $\uparrow$      & -           & 49.20 $\uparrow$      & 17.59 $\uparrow$      & 12.88  $\uparrow$     & 14.63       & -                    & 8.29        & 13.75  $\uparrow$      & 4.52        & 5.45  \\
\midrule
                                 & \multicolumn{12}{c}{\textbf{FC Samples}}   \\\midrule
\textbf{Qwen2-VL}             & 74.18        & 66.99       & 91.90       & 58.77       & 58.34       & 37.99       & 59.80       & 34.25       & 34.51       & 45.99        & 35.39       & 39.40  \\
\rowcolor{blue!10} \textbf{w \textsc{MCDGraph}}            & 85.24  $\uparrow$      & 66.42       & 90.93       & 81.77  $\uparrow$     & 77.30  $\uparrow$     & 56.79  $\uparrow$     & 76.47  $\uparrow$     & 53.06 $\uparrow$      & 52.85 $\uparrow$      & 59.45  $\uparrow$      & 58.51 $\uparrow$      & 58.72 $\uparrow$ \\
\textbf{InternVL2}             & 72.82        & 41.92       & 96.19       & 65.34       & 73.36       & 61.51       & 77.15       & 52.21       & 34.79       & 55.23        & 57.50       & 53.17      \\
\rowcolor{blue!10}\textbf{w \textsc{MCDGraph}}            & 73.86 $\uparrow$       & 72.70 $\uparrow$      & 94.28       & 59.99       & 78.23 $\uparrow$      & 69.02 $\uparrow$      & 74.32       & 52.12       & 54.71 $\uparrow$      & 57.41 $\uparrow$       & 68.51 $\uparrow$      & 44.20                 \\
\bottomrule
\end{tabular}}

        \caption{Model performance (Acc/F1/EM\_F1 for QA and F1 for FC) on various tasks with \textbf{Node-Edge} naming convention.}
            \label{tab:naming_conventions_node_edge}
\end{table*}

\begin{table*}[t]

   \centering
    \resizebox{0.95\linewidth}{!}{
\begin{tabular}{l|cccccccccccc}
\toprule
\multirow{2.5}{*}{\textbf{Models}} & \multicolumn{5}{c}{\textbf{Understanding}}                           & \multicolumn{7}{c}{\textbf{Reasoning}}                                                           \\  \cmidrule(lr){2-6}  \cmidrule(lr){7-13}
                                 & \textbf{NNu} & \textbf{EN} & \textbf{DC} & \textbf{DQ} & \textbf{NQ} & \textbf{NN} & \textbf{CR} & \textbf{CN} & \textbf{RA} & \textbf{SRN} & \textbf{NR} & \textbf{CP} \\
                                 \midrule
                                 & \multicolumn{12}{c}{\textbf{QA Samples}}         \\
                                \midrule
\textbf{Qwen2-VL}              & 21.00        & 22.90       & -           & 51.10       & 17.78       & 11.80       & 18.31       & -           & 11.50       & 15.33        & 8.66        & 14.02     \\
\rowcolor{blue!10}\textbf{w \textsc{MCDGraph}}            & 13.50        & 25.70 $\uparrow$      & -           & 63.30 $\uparrow$      & 18.66  $\uparrow$     & 9.56        & 18.91 $\uparrow$       & -           & 10.13       & 15.90 $\uparrow$       & 10.81  $\uparrow$     & 11.72  \\
\textbf{InternVL2}            & 32.80        & 17.70       & -           & 55.10       & 19.77       & 14.61       & 21.60       & -           & 10.69       & 16.20        & 11.29       & 11.79    \\
\rowcolor{blue!10}\textbf{w \textsc{MCDGraph}}           & 24.40        & 36.00 $\uparrow$      & -           & 54.20       & 21.10  $\uparrow$     & 13.32       & 21.44       & -           & 8.15        & 16.01        & 17.77  $\uparrow$     & 11.84 $\uparrow$  \\
\midrule
                                 & \multicolumn{12}{c}{\textbf{FC Samples}}   \\\midrule
\textbf{Qwen2-VL}             & 86.35        & 67.03       & 93.73       & 69.39       & 74.05       & 41.30       & 71.54       & 43.59       & 38.54       & 58.05        & 42.43       & 44.18   \\
\rowcolor{blue!10} \textbf{w \textsc{MCDGraph}}           & 68.68        & 69.10 $\uparrow$      & 96.45 $\uparrow$      & 78.20  $\uparrow$     & 76.96 $\uparrow$      & 57.61 $\uparrow$      & 82.66 $\uparrow$      & 56.07 $\uparrow$      & 55.19 $\uparrow$      & 63.51   $\uparrow$     & 63.17  $\uparrow$     & 56.45  $\uparrow$  \\
\textbf{InternVL2}           & 69.34        & 39.80       & 87.75       & 62.98       & 76.17       & 69.06       & 72.14       & 57.15       & 39.82       & 56.87        & 65.83       & 48.31         \\
\rowcolor{blue!10}\textbf{w \textsc{MCDGraph}}           & 78.05  $\uparrow$      & 71.65  $\uparrow$     & 75.85       & 55.84       & 81.23  $\uparrow$     & 71.50  $\uparrow$     & 79.29 $\uparrow$      & 52.43       & 57.76 $\uparrow$      & 53.85        & 67.60 $\uparrow$      & 41.99           \\
\bottomrule
\end{tabular}}

        \caption{Model performance (Acc/F1/EM\_F1 for QA and F1 for FC) on various tasks with \textbf{Name-R} naming convention.}
            \label{tab:naming_conventions_name_r}
\end{table*}

\begin{figure*}[t]
    \centering
    \includegraphics[width=\linewidth]{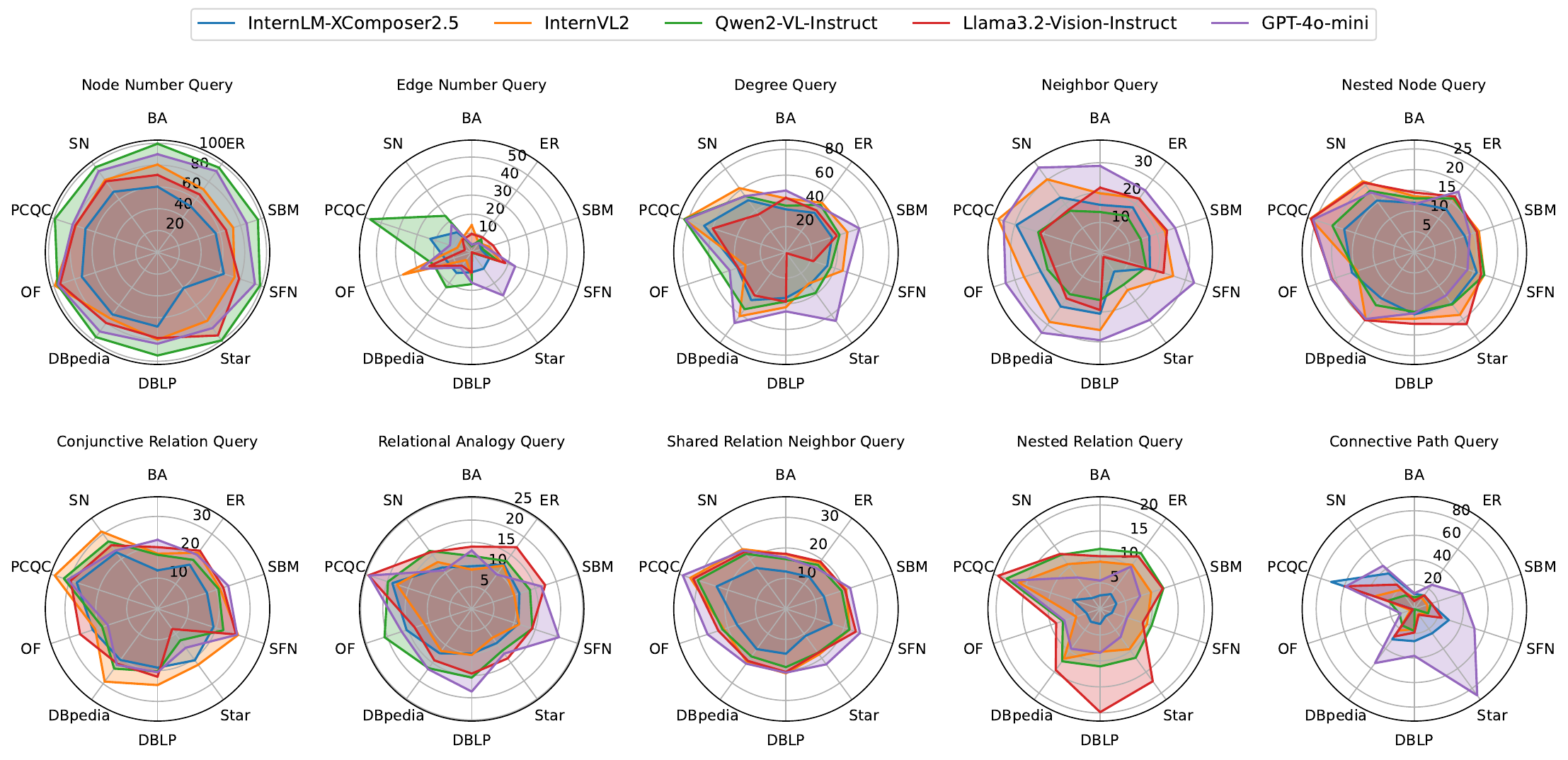}
    \vspace{-2em}
    \caption{Model performance (Acc/F1) on QA samples across various graph structures and tasks.}
    \label{fig:graph_results_qa}
    \vspace{-4em}
\end{figure*}

\begin{figure*}[t]
    \centering
    \includegraphics[width=\linewidth]{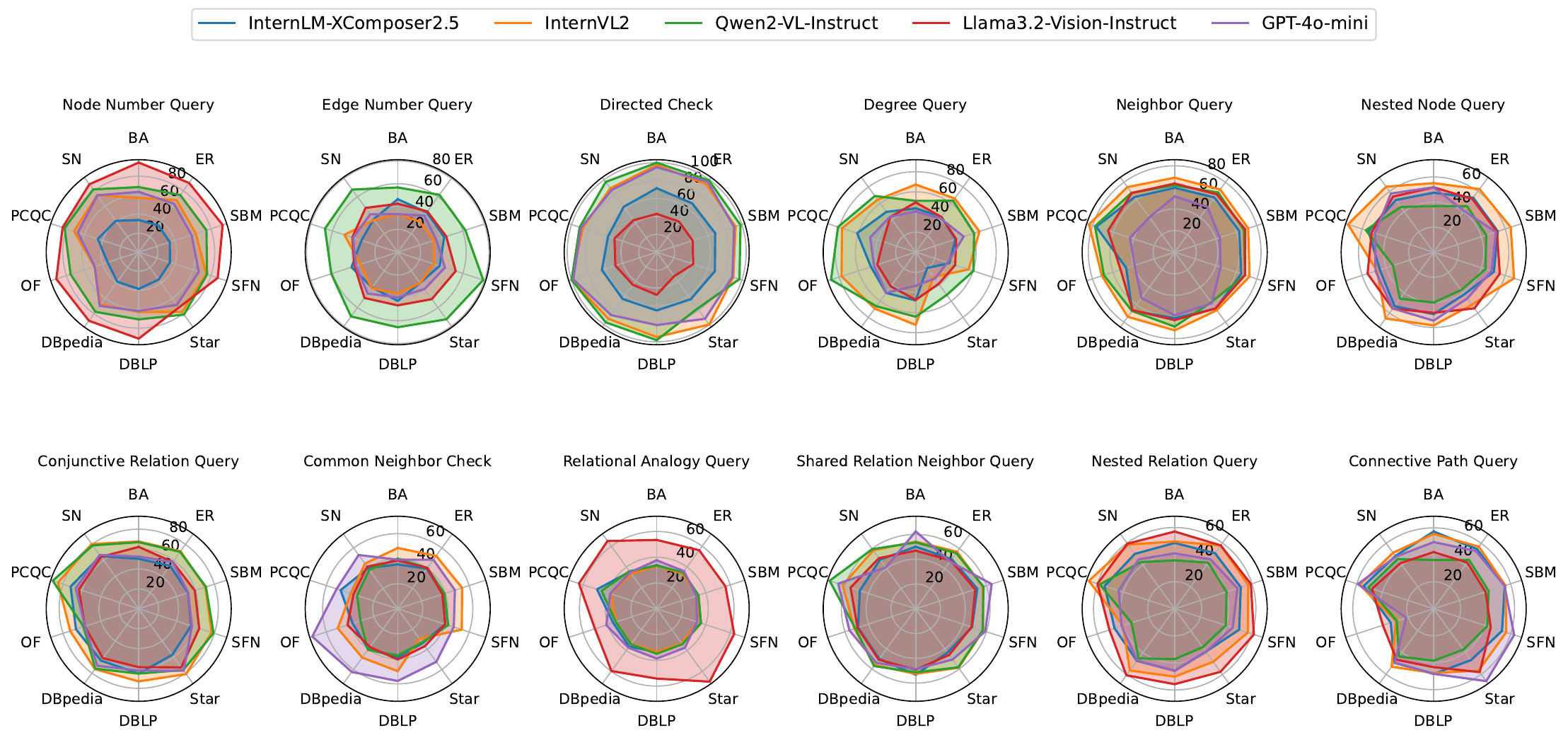}
    \vspace{-2em}
    \caption{Model performance (F1) on FC samples across various graph structures and tasks.}
    \label{fig:graph_results_fc}
    \vspace{-4em}
\end{figure*}

\begin{figure*}[t]
    \centering
    \includegraphics[width=\linewidth]{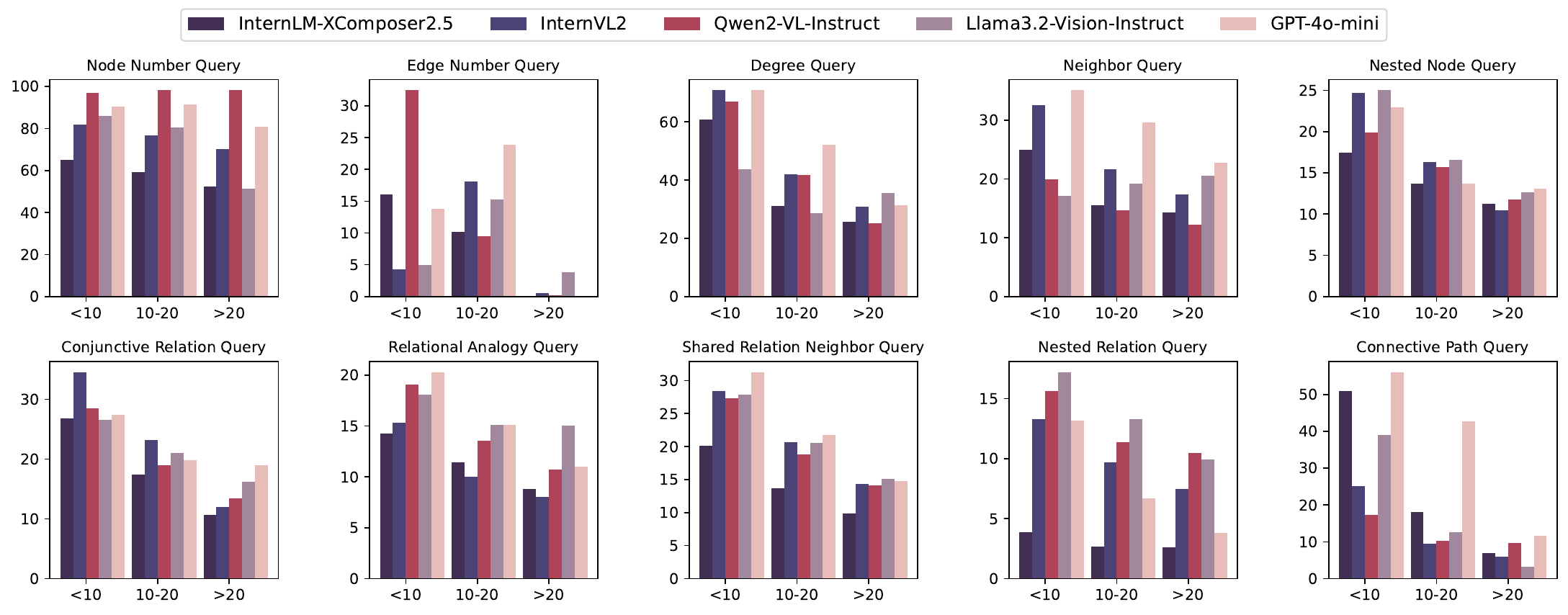}
    \vspace{-2em}
    \caption{Model performance (F1/Acc) comparison on QA samples across various tasks and \textbf{edge} ranges.}
    \label{fig:size_results_qa}
\end{figure*}

\begin{figure*}[t]
    \centering
    \includegraphics[width=\linewidth]{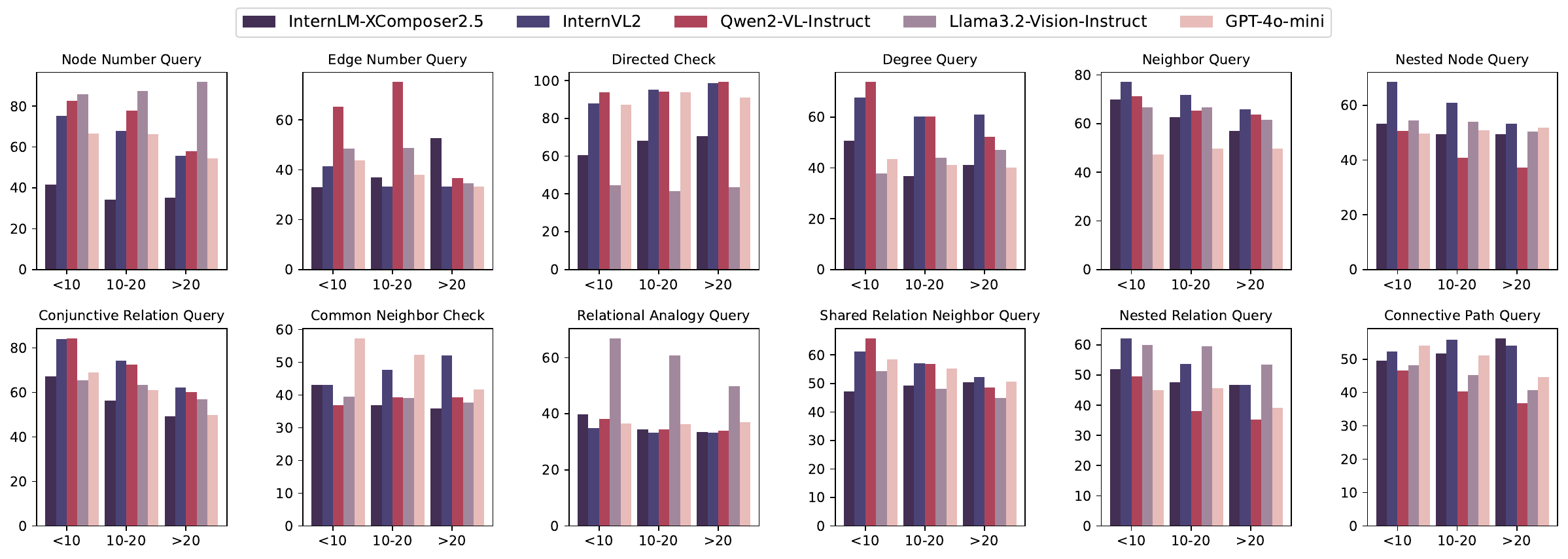}
    \caption{Model performance (F1) comparison on FC samples across various tasks and \textbf{edge} ranges.}
    \label{fig:size_results_fc}
\end{figure*}

\begin{figure*}[t]
    \centering
    \includegraphics[width=\linewidth]{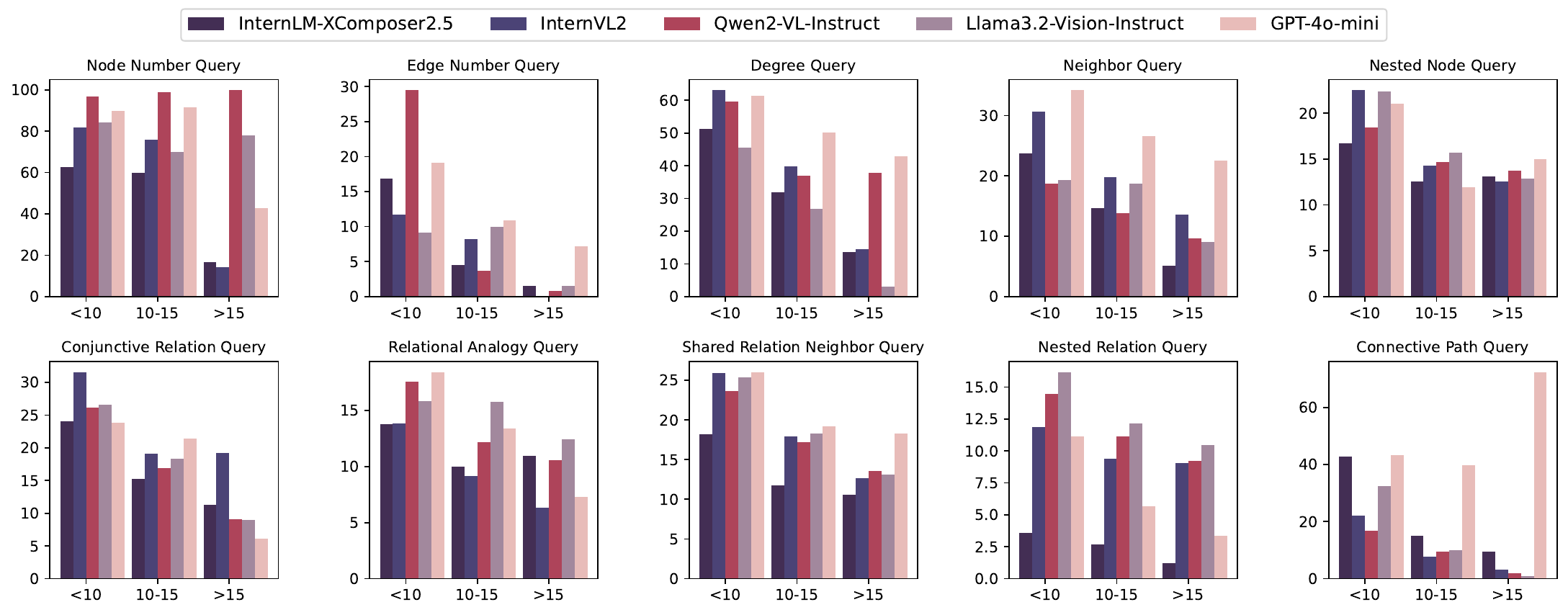}
    \caption{Model performance (F1/Acc) comparison on QA samples across various tasks and \textbf{node} ranges.}
    \label{fig:nodes_results_qa}
\end{figure*}

\begin{figure*}[t]
    \centering
    \includegraphics[width=\linewidth]{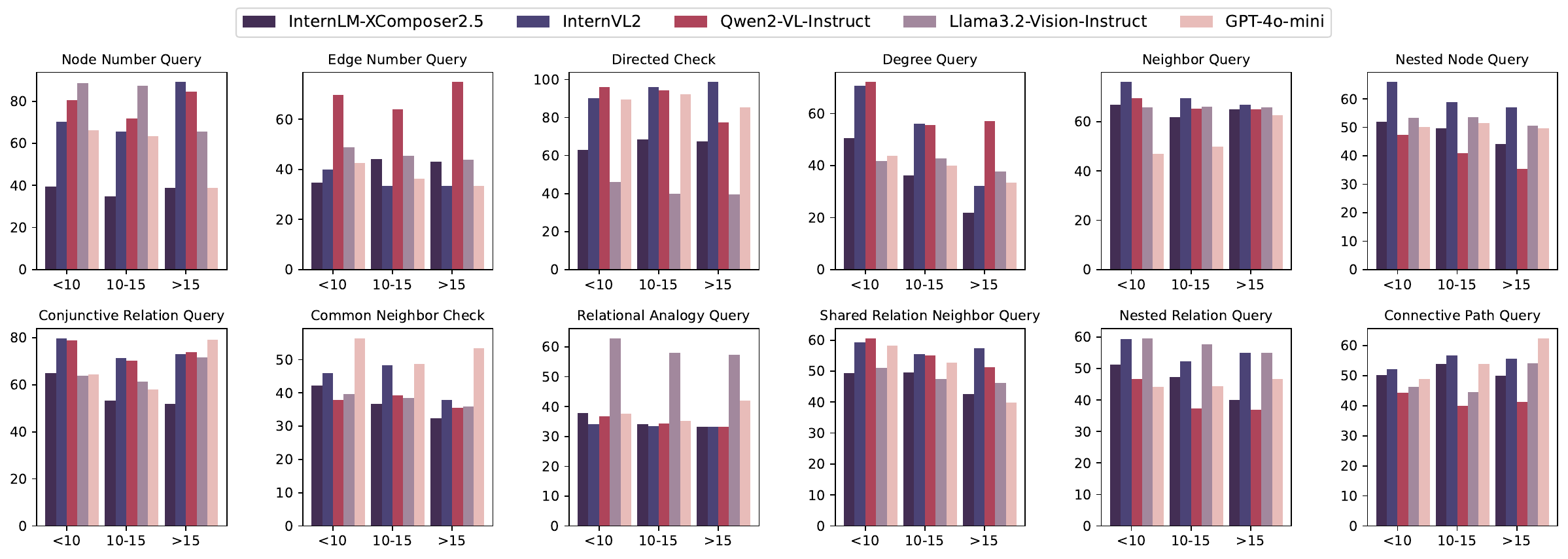}
    \caption{Model performance (F1) comparison on FC samples across various tasks and \textbf{node} ranges.}
    \label{fig:nodes_results_fc}
\end{figure*}

\begin{figure*}[t]
    \centering
    \includegraphics[width=\linewidth]{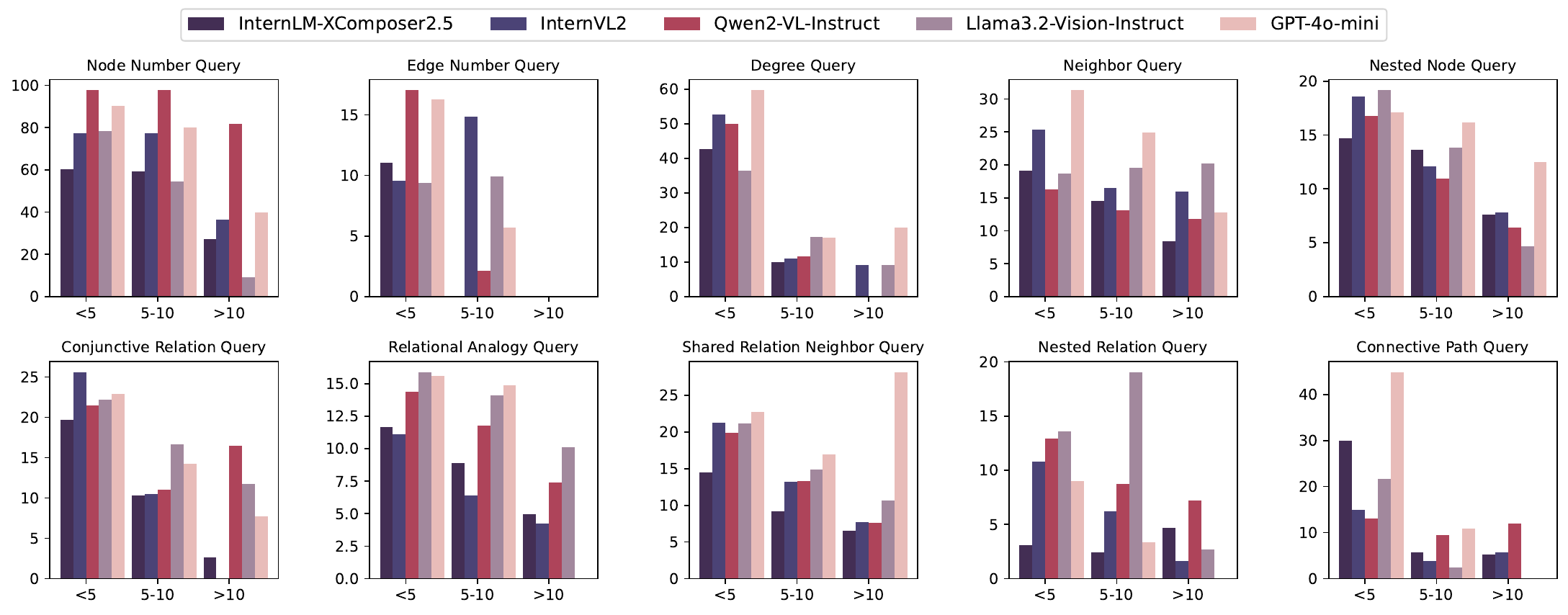}
    \caption{Model performance (F1/Acc) comparison on QA samples across various tasks and \textbf{average degree}.}
    \label{fig:degree_results_qa}
\end{figure*}

\begin{figure*}[ht]
    \centering
    \includegraphics[width=\linewidth]{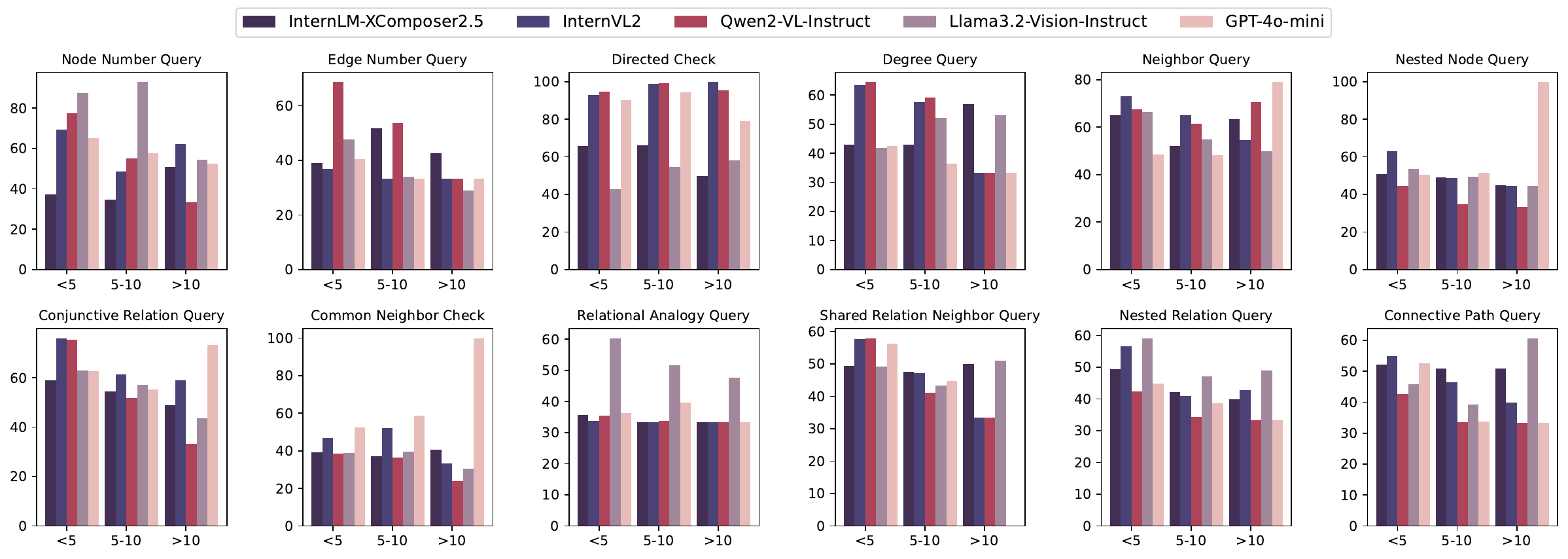}
    \caption{Model performance (F1) comparison on FC samples across various tasks and \textbf{average degree}.}
    \label{fig:degree_results_fc}
\end{figure*}

\begin{table*}[t]
    \centering
    \begin{minipage}{\textwidth}
    \centering
        \resizebox{\linewidth}{!}{
    \begin{tabular}{lccccccccccccccccc}
    \toprule
    \multicolumn{18}{c}{\textbf{QA Samples}} \\
    \toprule
    \multicolumn{1}{l}{\multirow{3.6}{*}{\textbf{Models}}}   & \multicolumn{5}{c}{\textbf{Understanding}}   & \multicolumn{12}{c}{\textbf{Reasoning}}                                                  \\
    \cmidrule(lr){2-6} \cmidrule(lr){7-18}
    \multicolumn{1}{c}{}                   & \textbf{NNu}   & \textbf{EN}    & \textbf{DQ}    & \multicolumn{2}{c}{\textbf{NQ}}                    & \multicolumn{2}{c}{\textbf{NN}}  & \multicolumn{2}{c}{\textbf{CR}}  & \multicolumn{2}{c}{\textbf{RA}}  & \multicolumn{2}{c}{\textbf{SRN}} & \multicolumn{2}{c}{\textbf{NR}}  & \multicolumn{2}{c}{\textbf{CP}}   \\
    \cmidrule(lr){2-2} \cmidrule(lr){3-3} \cmidrule(lr){4-4} \cmidrule(lr){5-6} \cmidrule(lr){7-8} \cmidrule(lr){9-10} \cmidrule(lr){11-12} \cmidrule(lr){13-14} \cmidrule(lr){15-16} \cmidrule(lr){17-18}
    \multicolumn{1}{c}{}            & \textbf{Acc}   & \textbf{Acc}   & \textbf{Acc}   & \textbf{F1}    & \textbf{Hits@1}                      & \textbf{F1}    & \textbf{Hits@1} & \textbf{F1}    & \textbf{Hits@1} & \textbf{F1}    & \textbf{Hits@1} & \textbf{F1}    & \textbf{Hits@1} & \textbf{F1}    & \textbf{Hits@1} & \textbf{EM\_F1} & \textbf{Label\_Acc} \\
    \midrule
    \textbf{Qwen2-VL}                        & 97.80        & 16.38        & 48.09        & 16.18        & 38.12            & 16.52        & 28.34            & 21.02        & 48.57            & 14.19        & 27.96            & 19.48         & 56.42            & 12.73        & 25.07            & 12.90           & 38.06               \\
    \rowcolor{blue!10} w \textsc{MCDGraph}                                           & 98.34 $\uparrow$       & 25.92  $\uparrow$       & 60.94  $\uparrow$       & 25.44  $\uparrow$       & 63.44 $\uparrow$            & 13.32        & 28.97 $\uparrow$            & 26.14 $\uparrow$        & 63.84   $\uparrow$          & 13.14        & 28.42   $\uparrow$          & 20.74   $\uparrow$       & 62.21    $\uparrow$         & 14.44  $\uparrow$       & 23.71  $\uparrow$           & 11.95           & 63.98 $\uparrow$               \\
    \textbf{InternVL2}                                & 77.45        & 9.78         & 50.75        & 25.01        & 68.58            & 18.30        & 30.82            & 24.87        & 59.31            & 10.83        & 17.12            & 20.72         & 59.99            & 10.58        & 18.97            & 14.53           & 43.97               \\
   \rowcolor{blue!10} w \textsc{MCDGraph}                                           & 95.68 $\uparrow$        & 40.45  $\uparrow$       & 54.78     $\uparrow$    & 28.80 $\uparrow$        & 72.72   $\uparrow$          & 19.43    $\uparrow$     & 27.86            & 28.53  $\uparrow$       & 67.61    $\uparrow$         & 11.67  $\uparrow$       & 19.81 $\uparrow$            & 22.34  $\uparrow$        & 61.67   $\uparrow$          & 16.50 $\uparrow$        & 40.21  $\uparrow$           & 12.76           & 25.23    \\          
    \bottomrule
    \end{tabular}}
        \caption{Performance Improvement of \textsc{MCDGraph} on QA samples across various tasks. $\uparrow$ indicates an improvement compared to the original model.}
    \label{tab:MCDGraph_results_qa_all}
\end{minipage}

     \vspace{0.5cm}
     
\begin{minipage}{\textwidth}
\centering
        \resizebox{\linewidth}{!}{
        \renewcommand\arraystretch{1.2}
    \begin{tabular}{lcccccccccccccccccccccccc}
    \toprule
    \multicolumn{25}{c}{\textbf{FC Samples}} \\
    \toprule
    \multirow{3.6}{*}{\textbf{Models}}                                                                                                 & \multicolumn{10}{c}{\textbf{Understanding}}      & \multicolumn{14}{c}{\textbf{Reasoning}}                                                                                                     \\
    \cmidrule(lr){2-11} \cmidrule(lr){12-25}
    & \multicolumn{2}{c}{\textbf{NNu}} & \multicolumn{2}{c}{\textbf{EN}} & \multicolumn{2}{c}{\textbf{DC}} & \multicolumn{2}{c}{\textbf{DQ}} & \multicolumn{2}{c}{\textbf{NQ}}
                                          & \multicolumn{2}{c}{\textbf{NN}} & \multicolumn{2}{c}{\textbf{CR}} & \multicolumn{2}{c}{\textbf{CN}} & \multicolumn{2}{c}{\textbf{RA}} & \multicolumn{2}{c}{\textbf{SRN}} & \multicolumn{2}{c}{\textbf{NR}} & \multicolumn{2}{c}{\textbf{CP}}                              \\
        \cmidrule(lr){2-3} \cmidrule(lr){4-5} \cmidrule(lr){6-7} \cmidrule(lr){8-9} \cmidrule(lr){10-11} \cmidrule(lr){12-13} \cmidrule(lr){14-15} \cmidrule(lr){16-17} \cmidrule(lr){18-19} \cmidrule(lr){20-21} \cmidrule(lr){22-23} \cmidrule(lr){24-25}
                                          & \textbf{F1}    & \textbf{Acc}   & \textbf{F1}    & \textbf{Acc}   & \textbf{F1}    & \textbf{Acc}   & \textbf{F1}    & \textbf{Acc}   & \textbf{F1}     & \textbf{Acc}   & \textbf{F1}    & \textbf{Acc}   & \textbf{F1}    & \textbf{Acc}   & \textbf{F1}     & \textbf{Acc}   & \textbf{F1}    & \textbf{Acc}   & \textbf{F1}    & \textbf{Acc}   & \textbf{F1}    & \textbf{Acc}   & \textbf{F1}    & \textbf{Acc}              \\
    
    \midrule
    \textbf{Qwen2-VL}    & 76.50          & 77.67           & 68.26          & 68.27          & 94.92          & 94.94          & 64.32          & 67.40          & 67.34          & 68.76          & 44.17          & 53.77          & 74.28          & 75.24          & 38.52          & 51.10          & 35.28          & 50.55          & 57.07          & 59.18           & 42.08          & 52.95          & 42.25          & 53.38                     \\
    \rowcolor{blue!10} w \textsc{MCDGraph}                       & 89.58  $\uparrow$         & 89.69    $\uparrow$        & 65.80          & 68.64  $\uparrow$         & 95.84  $\uparrow$         & 95.84  $\uparrow$         & 77.10    $\uparrow$       & 77.23   $\uparrow$        & 79.75   $\uparrow$        & 80.03   $\uparrow$        & 60.71 $\uparrow$          & 63.35   $\uparrow$        & 83.07   $\uparrow$        & 83.08 $\uparrow$          & 53.90  $\uparrow$         & 54.12 $\uparrow$           & 53.48 $\uparrow$          & 53.69   $\uparrow$        & 64.17 $\uparrow$          & 64.21 $\uparrow$           & 64.12   $\uparrow$        & 65.51 $\uparrow$          & 60.11  $\uparrow$         & 60.17  $\uparrow$                 \\
    \textbf{InternVL2}            & 68.63          & 71.18           & 36.82          & 50.23          & 93.17          & 93.18          & 63.28          & 63.84          & 72.81          & 73.27          & 62.46          & 63.62          & 75.37          & 76.18          & 47.12          & 50.40          & 33.70          & 50.14          & 56.98          & 57.42           & 55.94          & 58.42          & 54.71          & 54.71                         \\
    \rowcolor{blue!10}
     w \textsc{MCDGraph}                       & 76.55 $\uparrow$          & 77.71 $\uparrow$            & 71.98 $\uparrow$          & 72.02 $\uparrow$          & 90.04          & 90.50          & 56.83          & 58.53          & 80.98   $\uparrow$        & 81.89   $\uparrow$        & 73.14  $\uparrow$         & 73.98 $\uparrow$          & 80.23 $\uparrow$          & 80.69    $\uparrow$       & 52.09    $\uparrow$       & 52.99  $\uparrow$         & 52.81  $\uparrow$         & 57.08  $\uparrow$         & 59.07   $\uparrow$        & 60.07  $\uparrow$          & 69.03   $\uparrow$        & 70.24   $\uparrow$        & 45.82          & 49.27      \\
    \bottomrule
    \end{tabular}}
    \caption{Performance Improvement of \textsc{MCDGraph} on FC samples across various tasks. $\uparrow$ indicates an improvement compared to the original model.}
        \label{tab:MCDGraph_results_fc_all}
    \end{minipage}
\end{table*}

\begin{figure*}[t]
    \centering
    \includegraphics[width=\linewidth]{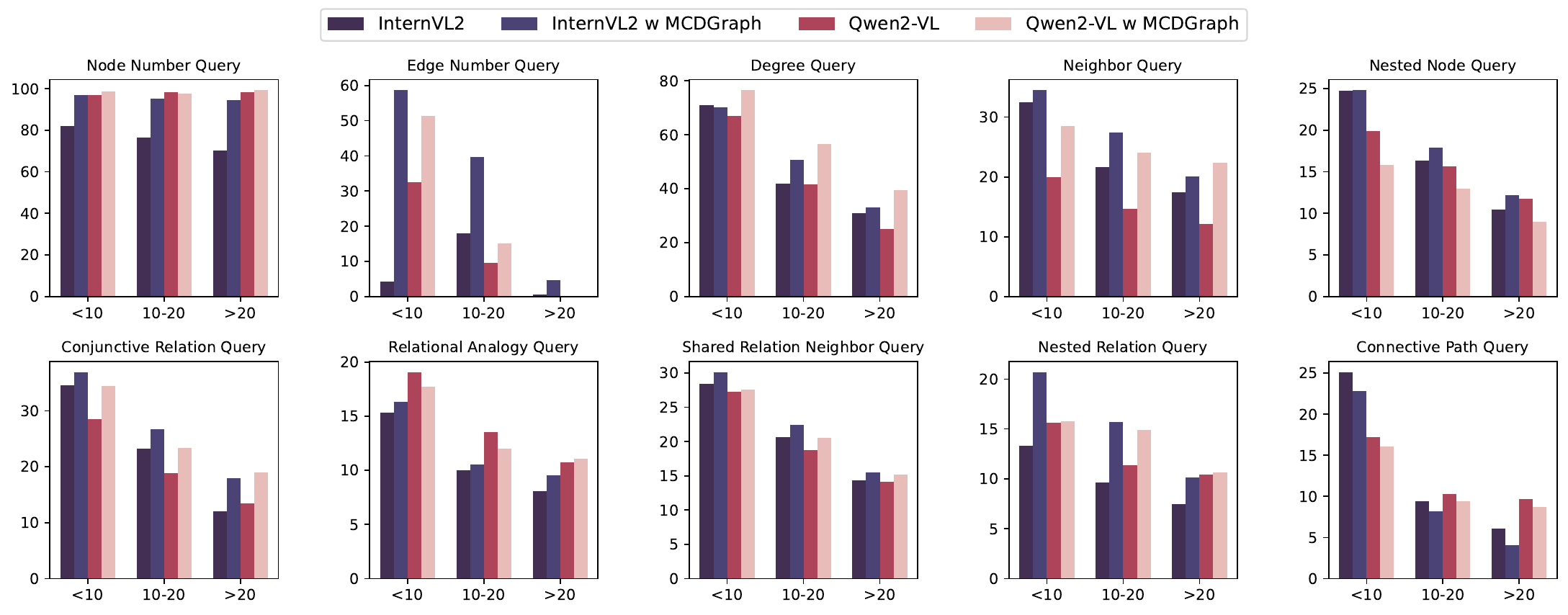}
    \caption{Performance Improvement of \textsc{MCDGraph} (F1/Acc) on QA samples across various tasks and \textbf{edge} ranges.}
    \label{fig:size_results_MCDGraph_qa}
\end{figure*}

\begin{figure*}[t]
    \centering
    \includegraphics[width=\linewidth]{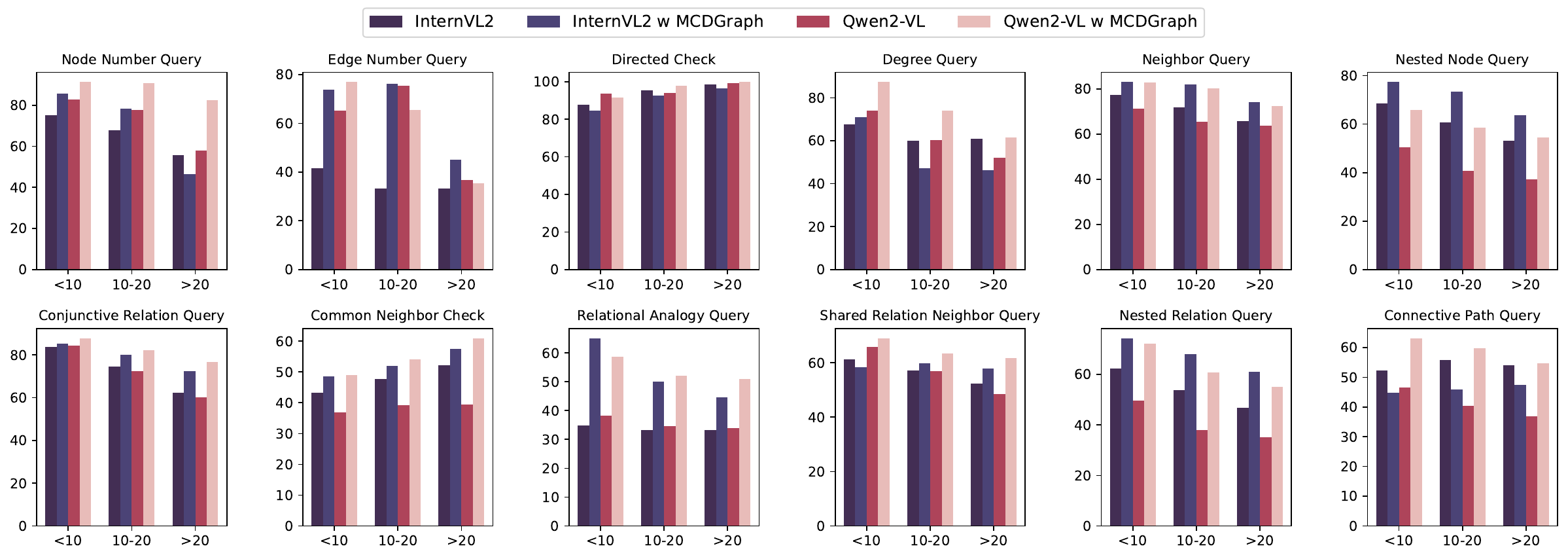}
    \caption{Performance Improvement of \textsc{MCDGraph} (F1) comparison on FC samples across various tasks and \textbf{edge} ranges.}
    \label{fig:size_results_MCDGraph_fc}
\end{figure*}

\begin{figure*}[t]
    \centering
    \includegraphics[width=\linewidth]{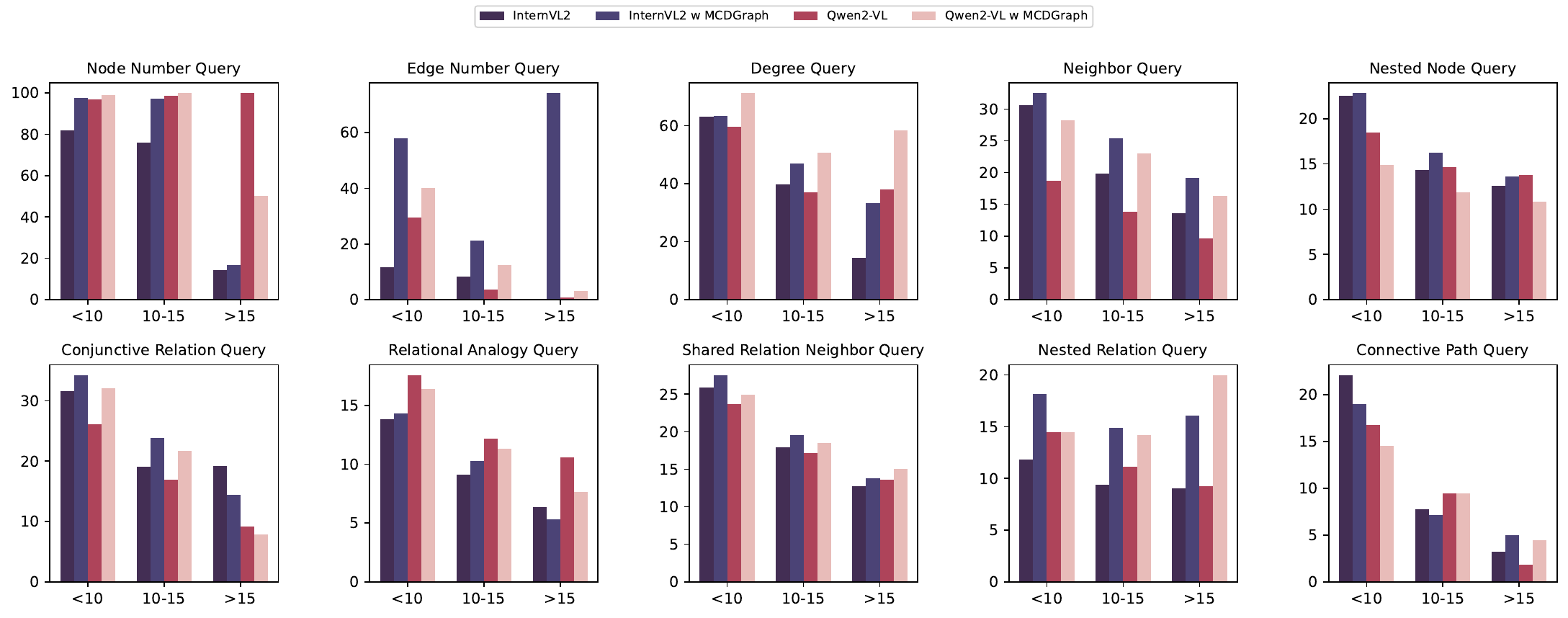}
    \caption{Performance Improvement of \textsc{MCDGraph} (F1/Acc) on QA samples across various tasks and \textbf{node} ranges.}
    \label{fig:nodes_results_MCDGraph_qa}
\end{figure*}

\begin{figure*}[t]
    \centering
    \includegraphics[width=\linewidth]{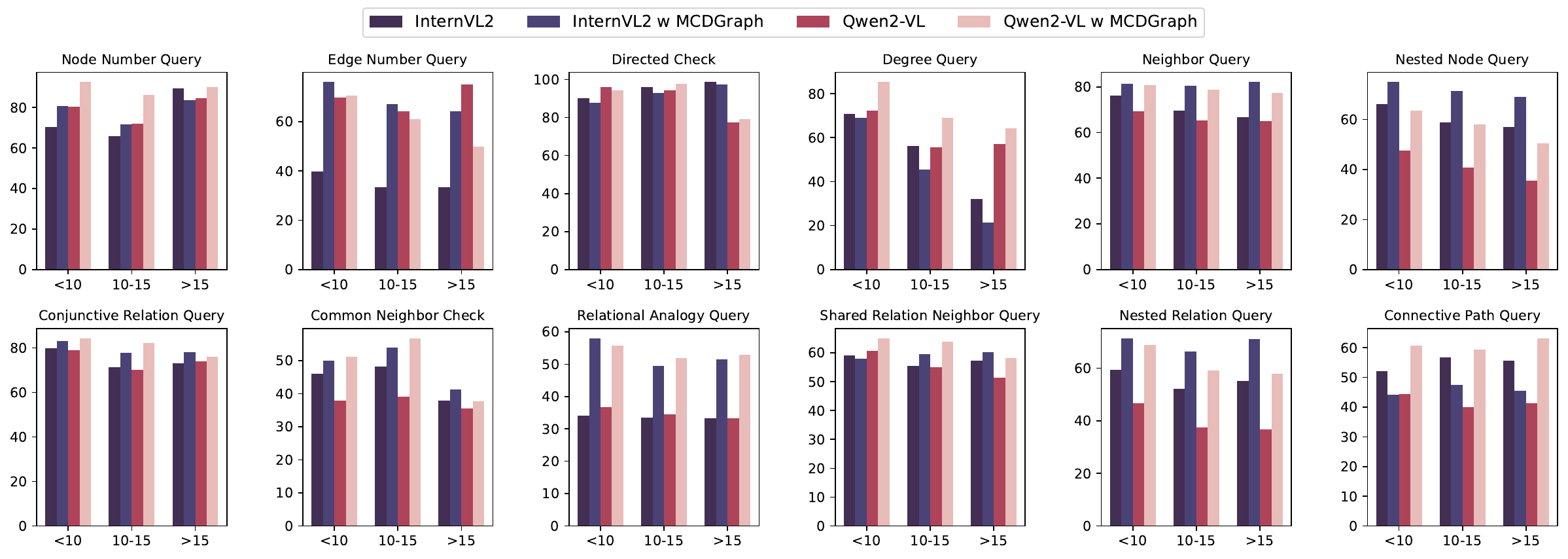}
    \caption{Performance Improvement of \textsc{MCDGraph} (F1) comparison on FC samples across various tasks and \textbf{node} ranges.}
    \label{fig:nodes_results_MCDGraph_fc}
\end{figure*}

\begin{figure*}[t]
    \centering
    \includegraphics[width=\linewidth]{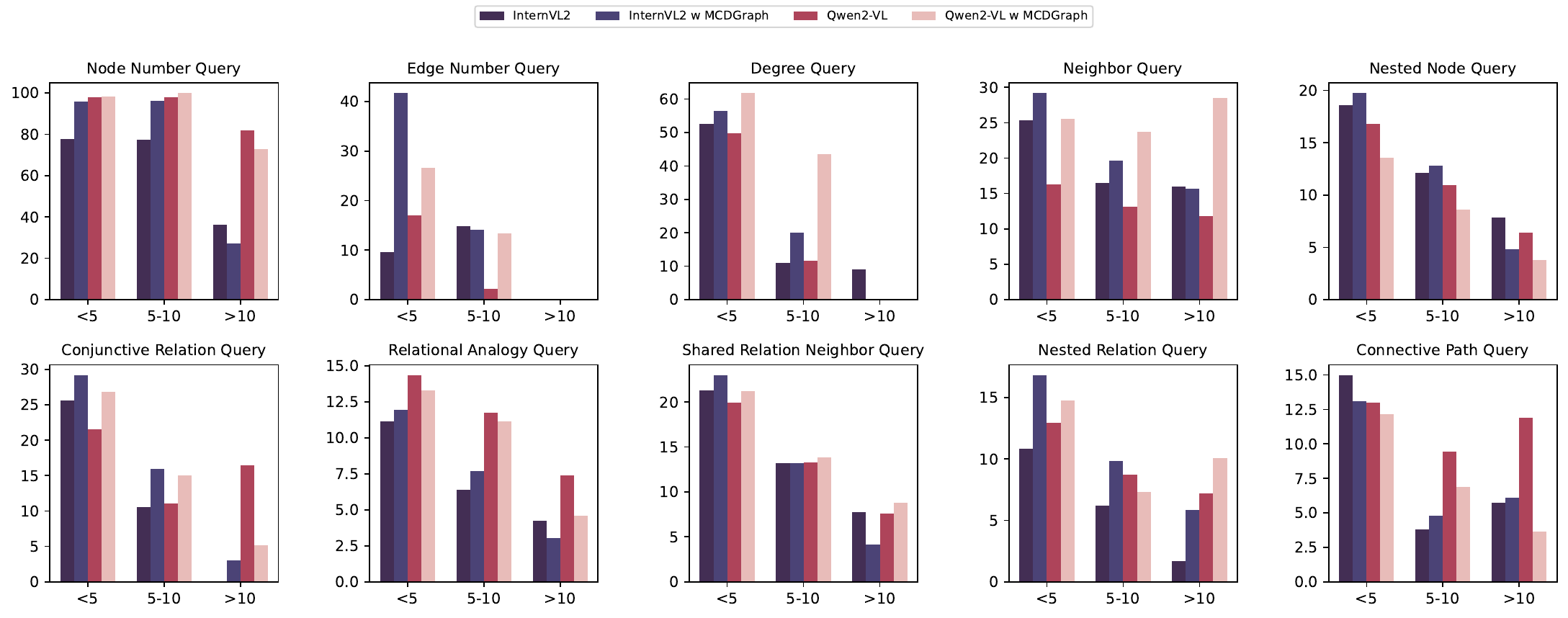}
    \caption{Performance Improvement of \textsc{MCDGraph} (F1/Acc) on QA samples across various tasks and \textbf{average degree}.}
    \label{fig:degree_results_MCDGraph_qa}
\end{figure*}

\begin{figure*}[t]
    \centering
    \includegraphics[width=\linewidth]{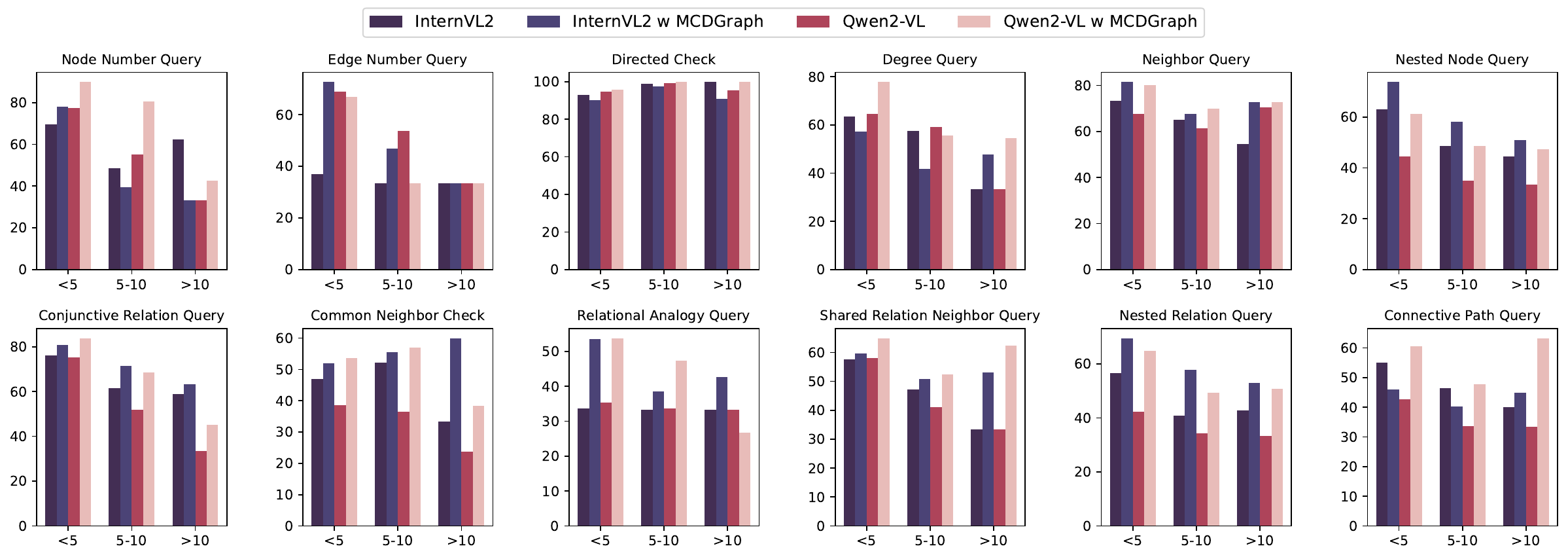}
    \caption{Performance Improvement of \textsc{MCDGraph} (F1) comparison on FC samples across various tasks and \textbf{average degree}.}
    \label{fig:degree_results_MCDGraph_fc}
\end{figure*}

\end{document}